\documentclass[11pt]{article}

\usepackage{acl}

\usepackage{times}
\usepackage{latexsym}

\usepackage[T1]{fontenc}

\usepackage[utf8]{inputenc}

\usepackage{microtype}

\usepackage{inconsolata}

\usepackage{graphicx}

\usepackage{algorithm}
\usepackage{enumitem}
\usepackage{newfloat}
\usepackage{listings}
\usepackage[most]{tcolorbox}
\usepackage{xcolor}  
\usepackage{booktabs}
\usepackage{multirow}
\usepackage{amsmath, amssymb}
\usepackage{booktabs}
\usepackage[table]{xcolor}
\usepackage{tabularx}
\usepackage{pifont}
\usepackage{algorithm}
\usepackage{algpseudocode} 
\usepackage{CJKutf8}
\usepackage{fontawesome5}
\usepackage{makecell}
\tcbuselibrary{breakable}
\usepackage[misc]{ifsym}

\definecolor{myred}{RGB}{207,62,62}
\hypersetup{
    colorlinks=true,            
    linkcolor=myred,              
    urlcolor=magenta            
}

\newcommand{\pmstd}[2]{%
  #1\ensuremath{\raisebox{0.15ex}{\tiny$\pm$#2}}%
}

\title{ProMed: Shapley Information Gain Guided \\Reinforcement Learning for Proactive Medical LLMs}

\author{
  Hongxin Ding\textsuperscript{1,2,3}\footnotemark[1], 
  Baixiang Huang\textsuperscript{1,2,3}\footnotemark[1], 
  Yue Fang\textsuperscript{1,2,3}\footnotemark[1], 
  Weibin Liao\textsuperscript{1,2,3}\footnotemark[1],\\
  \textbf{Xinke Jiang}\textsuperscript{1,2,3}, 
  \textbf{Jinyang Zhang}\textsuperscript{1,2,3}, 
  \textbf{Yinghao Zhu}\textsuperscript{5},
  \textbf{Zheng Li}\textsuperscript{2},\\
  \textbf{Liantao Ma}\textsuperscript{1,3,4},
  \textbf{Junfeng Zhao}\textsuperscript{2,3}\footnotemark[2], 
  \textbf{Yasha Wang}\textsuperscript{1,3,4}\footnotemark[2]\\
  \textsuperscript{1}National Engineering Research Center of Software Engineering, Peking University, China \\
  \textsuperscript{2}School of Computer Science, Peking University, Beijing, China \\
  \textsuperscript{3}Key Laboratory of High Confidence Software Technologies, Ministry of Education \\
  \textsuperscript{4}Peking University Information Technology Institute, Tianjin Binhai, China \\
  \textsuperscript{5}School of Computing and Data Science, The University of Hong Kong\\
  \textrm{\Letter}~\texttt{\{dinghx, zhaojf, wangyasha\}@pku.edu.cn}
}

\newcommand{\M}{\text{ProMed}}

\begin{document}
\maketitle
\begin{abstract}
Interactive medical questioning is essential in clinical consultations, where physicians must actively gather necessary patient information. Yet existing medical Large Language Models (LLMs) predominantly follow a reactive paradigm, risking diagnostic errors by answering before seeking sufficient details. To bridge this gap, we propose \textbf{\M}, a reinforcement learning framework that transitions LLMs toward a proactive paradigm, enabling them to ask clinically valuable questions before decision-making. Central to \M~is the Shapley Information Gain (SIG) reward, which quantifies a question's clinical utility as the amount of newly acquired information, while considering its contextual importance via Shapley values. We integrate SIG into a two-stage training pipeline: (1) SIG-Guided Model Initialization uses Monte Carlo Tree Search to construct high-reward interaction trajectories for supervision, and (2) SIG-Augmented Policy Optimization, with a novel SIG-guided Reward Distribution Mechanism that prioritizes informative questions for fine-grained optimization. Experiments on partial-information medical benchmarks show that \M~significantly outperforms state-of-the-art methods by 6.29\% on average and delivers a 54.45\% gain over the reactive paradigm, and generalizes robustly to out-of-domain cases. Our codes are available at {\url{https://github.com/hxxding/ProMed}}.
\end{abstract}

\begingroup
\renewcommand\thefootnote{}\footnotetext{$\ast$~These authors contribute equally.}
\renewcommand\thefootnote{}\footnotetext{$\dagger$~Corresponding authors.}
\addtocounter{footnote}{-1}
\endgroup

\section{Introduction}

\begin{figure}[!t]
    \centering
    \includegraphics[width=1\linewidth]{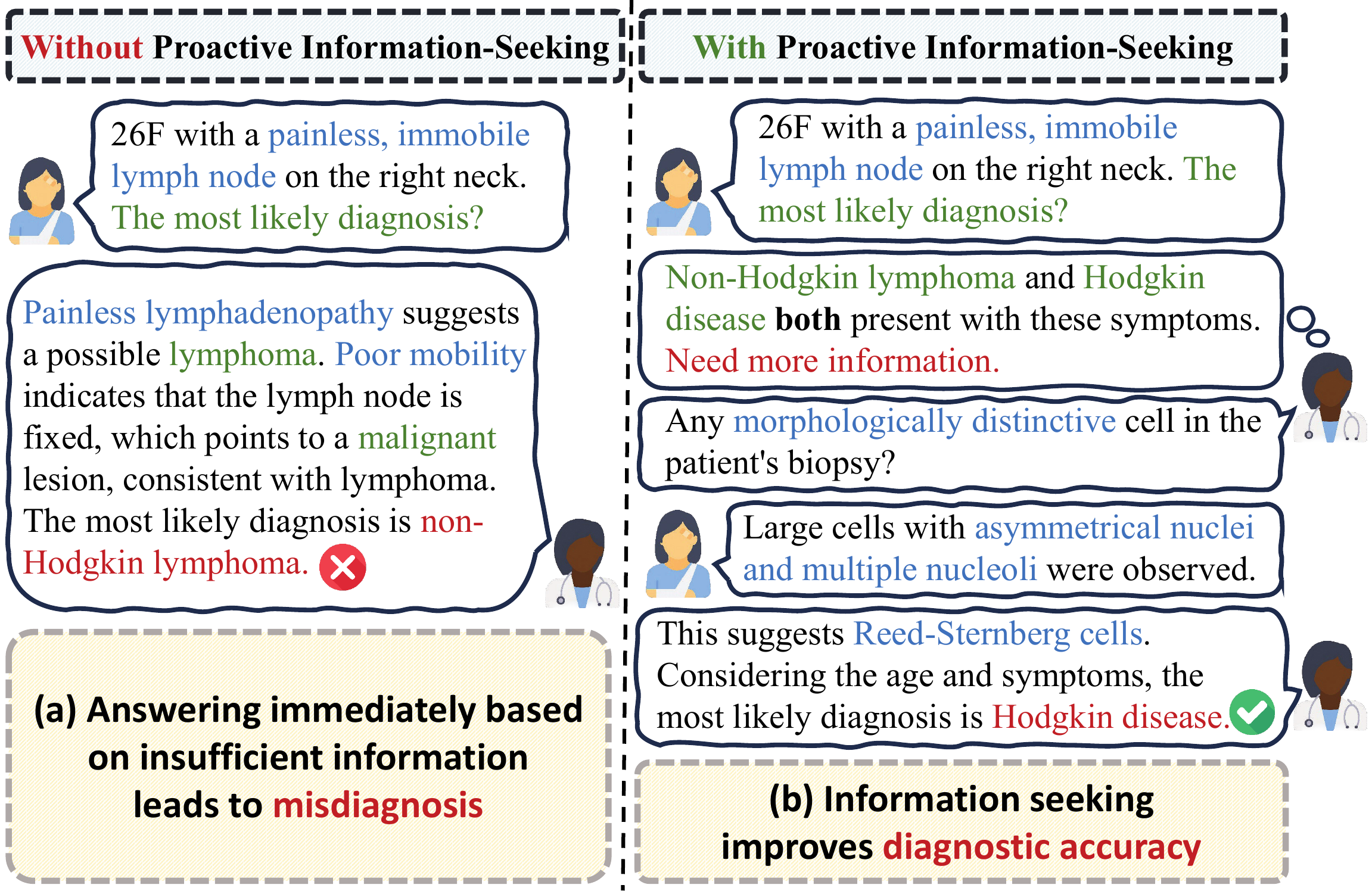}
    \caption{Clinical consultation example: relying on partial information causes misdiagnosis (a), actively seeking information enables accurate diagnosis (b).}
    \label{fig:fig1}
\vspace{-0.15cm}
\end{figure}

Clinical diagnosis fundamentally relies on interactive medical consultation. Patients typically initiate a clinical encounter with vague or incomplete chief complaints, requiring clinicians to proactively elicit relevant historical and symptomatic information through targeted questioning before an accurate diagnosis can be established.
Recently, leveraging large language models (LLMs)~\cite{singhal2025medPalm,wu2024pmc-llama,zhang2023huatuogpt} for clinical decision-making has emerged as an active research direction. Benefiting from large-scale pretraining on medical knowledge, LLMs have demonstrated promising performance on static clinical tasks, such as medical examinations~\cite{ding20243ds,jiang-etal-2025-hykge} and disease diagnosis~\cite{mcduff2025diagnosis,xu2025dearllm}.
However, when deployed in interactive clinical settings, existing LLMs remain constrained by a \textbf{reactive paradigm} that generates predictions solely based on an initial query. In such cases, partial or biased information in the initial patient input can induce cognitive bias in the model’s reasoning process, leading to erroneous clinical decisions, potential misdiagnosis, and compromised patient safety (as illustrated in Figure~\ref{fig:fig1}).
Therefore, \textit{enabling LLMs to transition from a \textbf{reactive paradigm} to a \textbf{proactive paradigm}, where models can systematically acquire clinically informative evidence through interaction and inquiry, has become a critical research problem for unbiased and reliable clinical decision-making.}


Recent efforts on \textbf{proactive paradigm} for interactive medical LLMs primarily rely on prompt engineering or supervised fine-tuning (SFT).
Despite their demonstrated effectiveness, prompt-based approaches~\cite{li2024mediq,hu2024UoT,liu2025MMD-eval,wang2025healthq,zhu2025ask-patient} merely induce questioning behaviors through carefully designed prompts. Through this approach, LLMs passively follow question-asking instructions, rather than genuinely transitioning to an active reasoning paradigm, resulting in limited performance gains.
In contrast, SFT-based methods~\cite{liu2025dialogueT,liao2023automatic_consultation} attempt to simulate interactions by training on static multi-turn dialogues. However, such approaches lack robustness and adaptability to diverse and unpredictable patient scenarios encountered in real-world clinical practice. More critically, empirical studies~\cite{liao2025tpo,hong2024orpo,liao2025learnat} have shown that while SFT relying solely on positive feedback encourages models to ask clinically relevant questions, it also inadvertently amplifies incorrect or misleading inquiries.
Therefore, \textit{enabling a genuine \textbf{paradigm shift}, one that \textbf{promotes the generation of clinically informative questions} while \textbf{restraining erroneous or misleading inquiries}, remains a fundamental problem in interactive medical consultation.}


Reinforcement learning (RL) offers a promising solution for achieving this goal, as its \textbf{reward mechanism} enables explicit encouragement or suppression of model behaviors.
Within this context, we identify \textbf{two key challenges} in applying RL to proactive interactive medical consultation.
\begin{itemize}
[leftmargin=*,itemsep=0pt,parsep=0.5em,topsep=0.3em,partopsep=0.3em]
    \item \textbf{Reward Modeling.} Quantifying the clinical value of a question has long been a central problem in interactive medical diagnosis. Existing studies typically rely on heuristic strategies, such as LLM–based scoring~\cite{wang2025healthq} or leave-one-out evaluations that assess a question’s impact on model confidence~\cite{hu2024UoT,lee2025good-utility-estimation,mazzaccara2024EIG,zhu2025ask-patient}. However, these approaches overlook a critical property of medical reasoning: its inherently \textbf{compositional nature}. Accurate diagnosis often depends on joint considerations of multiple clinical facts, and the utility of a single question may only emerge when combined with others. Consequently, isolated evaluation of individual questions can be fundamentally insufficient.
    \item \textbf{Reward Attribution.} Assigning terminal rewards to individual tokens along a trajectory is a long-standing challenge in RL. Naïve strategies such as uniformly distributing rewards across all tokens in a trajectory (e.g., GRPO~\cite{shao2024deepseekmath-grpo}), \textbf{fail to distinguish clinically salient questions} from irrelevant ones, assigning them equal credits. Moreover, token-level averaging implicitly biases the model toward generating longer questions, as longer sequences receive more total rewards, thereby encouraging verbosity rather than clinical informativeness.
\end{itemize}

To address these challenges, we propose an RL framework for training \textbf{\underline{Pro}}active \textbf{\underline{Med}}ical LLMs (\textbf{\M}). \textbf{For Reward Modeling}, we introduce the novel \textit{Shapley Information Gain (SIG)} reward mechanism. SIG utilizes Shapley values~\cite{winter2002shapley} from cooperative game theory to measure the importance of medical information while considering its interactions, yielding a context-aware information gain to precisely quantify questions' clinical utility. For \textbf{Reward Attribution}, we closely integrate SIG into RL through a two-stage design. \textit{Stage 1: SIG-Guided Model Initialization} employs Monte Carlo Tree Search (MCTS) with SIG rewards to systematically explore optimal doctor-patient interaction trajectories for supervised warm-up, improving stability and convergence for weak initial policies~\cite{wang2025beyond,xu2025kdrl} and alleviating the scarcity of high-quality medical interaction data. \textit{Stage 2: SIG-Augmented Policy Optimization} incorporates SIG into GRPO with a novel \textit{SIG-Guided Reward Distribution Mechanism}. Unlike standard GRPO that assigns uniform rewards to all tokens, our strategy allocates rewards proportionally to questions’ utility, enabling more targeted, fine-grained policy optimization that reinforces the LLM's proactive ability.

\noindent\textbf{Our contributions are as follows}:
\begin{itemize}[leftmargin=*,noitemsep,topsep=2pt]
    \item \textbf{Insightfully}, we pioneer an RL framework \M~to shift medical LLMs from reactive responders to proactive information seekers.
    \item \textbf{Technically}, we develop the SIG reward that leverages Shapley to model medical information interactions and quantify question utility, with a tailored reward distribution mechanism for fine-grained optimization.
    \item \textbf{Experimentally}, extensive evaluations demonstrate that \M~significantly outperforms existing methods and exhibits robust generalization to out-of-distribution (OOD) benchmarks.
    \item \textbf{Practically}, we construct two public benchmarks targeting interactive medical questioning with standardized splits to facilitate future research.
\end{itemize}

\section{Preliminaries}
\label{sec:task_definition}

We formulate the \textbf{Interactive Medical Questioning} task, which mirrors realistic clinical consultations where patients provide incomplete information during initial inquiries. Each patient case in the dataset $\mathcal{D} = \{\mathcal{X}_i\}_{i=1}^N$ is defined as $\mathcal{X} = \{Q, \mathcal{F}, A^*\}$ and consists of:
\begin{itemize}[leftmargin=*,noitemsep,topsep=2pt]
    \item $\mathcal{F} = \{f_1, f_2, \dots, f_n\}$: complete set of atomic facts fully describing the patient's clinical condition, where each $f_i$ is a minimal, self-contained information unit (e.g., symptom or lab result);
    \item $Q$: atomic clinical inquiry without information;
    \item $A^*$: ground-truth answer based on the full set $\mathcal{F}$.
\end{itemize}

We model the LLM as an \textbf{interactive agent} that proactively acquires information through multi-turn questioning. The interaction starts with a \textbf{partial information question} $Q_p = (F_p, Q)$, where $F_p \subset \mathcal{F}$ represents limited patient information (e.g., a chief complaint). At each turn $t$, the model updates its internal belief state $s_{t-1}$ about the patient based on the current dialogue history $\mathcal{H}_{t-1} = {(q_1, r_1), \dots, (q_{t-1}, r_{t-1})}$. It decides whether information is sufficient and takes action $a_t$: either asking a \textbf{follow-up question} $q_t$ and receiving the patient response $r_t$, or terminating the interaction by outputting an answer $A'$.

\section{Methodology}

\subsection{Overview}
\begin{figure*}[ht]
    \centering
    \includegraphics[width=\linewidth]{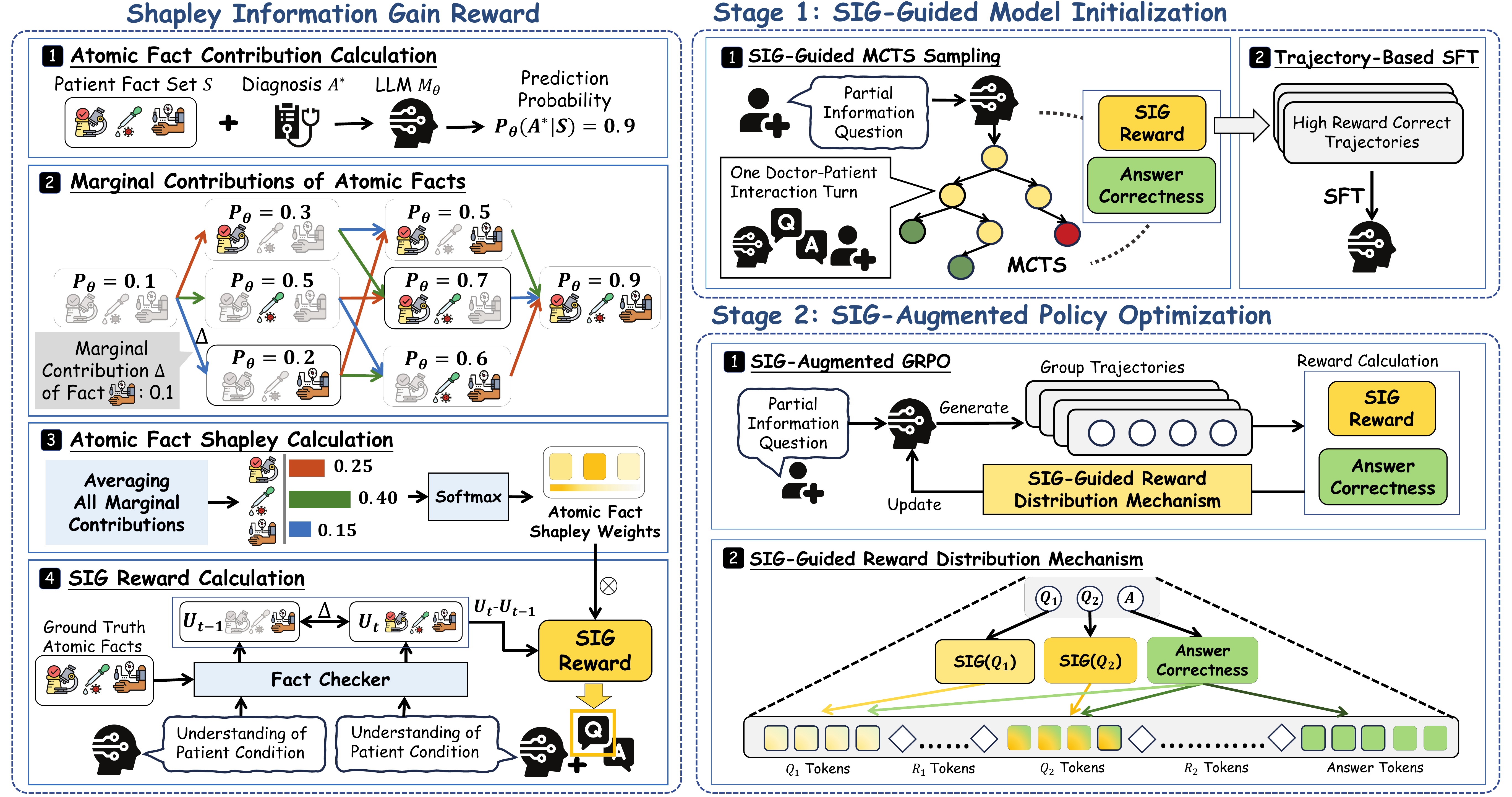}
    \caption{\textbf{\M~framework.} \textbf{Shapley Information Gain Reward} calculates rewards for questions. \textbf{Stage 1} generates high-reward trajectories via MCTS for SFT. \textbf{Stage 2} distributes rewards and optimizes the policy via RL. }
    \label{fig:framework_overview}
\end{figure*}

As illustrated in Figure~\ref{fig:framework_overview}, \M~comprises three modules: \textbf{Shapley Information Gain Reward} quantifies the clinical utility of questions to guide training. \textbf{SIG-Guided Model Initialization} uses MCTS with SIG to explore high-quality interaction trajectories for SFT. \textbf{SIG-Augmented Policy Optimization} integrates SIG into GRPO with a reward distribution for fine-grained optimization.

\subsection{Shapley Information Gain Reward}
\label{sec:SIG_Reward}
To guide LLMs' question-asking with accurate and clinical-aware rewards, we propose \textit{Shapley Information Gain} (SIG), which quantifies question utility by measuring newly acquired information amount, while accounting for fact importance and interactions via cooperative game theory.

\noindent\textbf{Atomic Fact Foundation.}
We measure \emph{incremental information} elicited by each question as the number of newly acquired atomic facts. Specifically, we leverage the pre-constructed ground-truth atomic fact set $\mathcal{F} = {f_1, f_2, \dots, f_n}$ defined in Section~\ref{sec:task_definition}. This enables explicit tracking of acquired facts and per-question information gain.

\noindent\textbf{State Approximation via Dynamic Understanding Generation.}
At dialogue turn $t$, after the model poses a follow-up question $q_t$ and receives the patient response $r_t$, the dialogue history is updated to:
$\mathcal{H}_t = \{(q_1, r_1), (q_2, r_2), \dots, (q_t, r_t)\}$. We approximate the model’s internal belief state $s_t$ using a \textit{doctor understanding prompt} (Appendix~\ref{appendix:prompts}) that instructs the model to articulate its understanding of patient condition $U_t$ based on $Q_p$ and $\mathcal{H}_t$. $U_t$ captures the model’s current grasp of patient information and serves as a proxy for $s_t$.

\noindent\textbf{Fact-Level Information Gain.}
To quantify the informational value of question $q_t$, we measure the incremental change in model understanding following its response $r_t$. Using a high-capacity LLM-powered fact-checker to determine fact entailment ($f_i \in \mathcal{F}$) in $U_t$, raw Information Gain (IG) is defined as the increase in fact coverage between current and previous model understanding states:
\begin{equation}
\small
IG(q_t) =\frac{1}{|\mathcal{F}|} \sum_{f_i \in \mathcal{F}} \left[ \mathbf{1}(f_i \subseteq U_t) - \mathbf{1}(f_i \subseteq U_{t-1}) \right]
\end{equation}
where $\mathbf{1}(\cdot)$ is the indicator function based on the fact-checker's judgments. The IG score captures the number of newly acquired facts elicited by $q_t$, but assumes equal and independent contributions across all facts.

\noindent\textbf{Atomic Fact Shapley Calculation.}
In clinical practice, information differs in diagnostic value and exhibits nontrivial interactions. For instance, a chest CT scan is typically more informative for pneumonia than a reported fever\cite{balafar2024chestCTcomparison}. The straightforward recall-based IG fails to capture such distinctions.
Traditional ``leave-one-out'' evaluations~\cite{hu2024UoT,zhu2025ask-patient} measure information importance by their inflicted change in model uncertainty, which treat facts in isolation and overlook their complex dependencies and interactions. For instance, diagnosing acute appendicitis relies on combined evidence: elevated white blood cell count, right lower abdominal tenderness, and Blumberg’s sign~\cite{snyder2018acuteappendicitis}. Omitting one symptom may not significantly affect the model's prediction if others are absent, thereby underestimating its true clinical value.
To capture both \textbf{(1) varying clinical importance} and \textbf{(2)complex interactions} of facts, we adopt the \textbf{Shapley value}~\cite{winter2002shapley} from cooperative game theory to more precisely and robustly attribute information importance.


Formally, given LLM $M_{\theta}$ parameterized by $\theta$, atomic question $Q$ and the desired answer $A^*$, the value of an atomic fact subset $S \subseteq \mathcal{F}$ is defined as the log-probability of predicting $A^*$ based on $Q$ and $S$:
\begin{equation}
\small
v(S) = \log P_{\theta}(A^* \mid Q,S)
\end{equation}
The value function quantifies the facts' contribution to the model's correct prediction, thus reflecting their utility. The Shapley value $\phi(f_i)$ of fact $f_i$ is the expected marginal gain in $v(S)$ when $f_i$ is added to all subsets:

\begin{equation}
\label{eq:fact_shapley}
\small
\phi(f_i) = \sum_{S \subseteq \mathcal{F} \setminus \{f_i\}} \frac{1}{|\mathcal{F}| \binom{|\mathcal{F}|-1}{|S|}} \left[ v(S \cup \{f_i\}) - v(S) \right]
\end{equation}
This calculation captures individual fact importance via marginal contributions, and their complex interactions by considering all possible fact combinations, thus addressing the two aforementioned clinical factors.

Since enumerating $2^{|\mathcal{F}|}$ subsets is computationally infeasible, we employ a \textit{Monte Carlo approximation} with online averaging. At iteration $k$, we sample a random permutation $\pi_k$ of $\mathcal{F}$, compute the marginal contribution of each fact then update its Shapley estimation:
\begin{equation}
\small
\phi^{(k)}(f_i) = \frac{k-1}{k} \phi^{(k-1)}(f_i) + \frac{1}{k} [v(S_i^{\pi_k}\cup f_i)-v(S_i^{\pi_k})].
\end{equation}
where $S_i^{\pi_k}$ denotes the set of facts preceding $f_i$ in $\pi_k$. Crucially, we batch all evaluations within a single permutation into one model pass, leveraging parallel inference to significantly accelerate computation. The process terminates when estimates converge within a tolerance $\epsilon$, enabling a controllable trade-off between efficiency and accuracy. Pseudo-codes and complexity analysis are provided in Appendix~\ref{appendix:monte_carlo_shapley}.

\noindent\textbf{Shapley Information Gain Reward Calculation.}
Once the Shapley values $\{\phi(f_1), \phi(f_2), \dots, \phi(f_n)\}$ are obtained, we compute softmax-normalized weights:
\begin{equation}
\small
\tilde{\phi}_i = \frac{\exp(\phi(f_i))}{\sum_{j=1}^{n}\exp(\phi(f_j))},
\end{equation}
Shapley Information Gain (SIG) for question $q_t$ is:
\begin{equation}
\small
\label{eq:sig_reward}
\text{SIG}(q_t) = \sum_{f_i \in \mathcal{F}} \tilde{\phi}_i  \left[ \mathbf{1}(f_i \subseteq U_t) - \mathbf{1}(f_i \subseteq U_{t-1}) \right],
\end{equation}
This formulation captures the importance-weighted information gain induced by a question, encouraging the model to prioritize acquiring information that is both novel and clinically impactful. It can be used to guide SFT data collection and drive policy optimization via reinforcement learning.

\subsection{SIG-Guided Model Initialization}



This stage initializes the LLM's information-seeking policy on high-quality interactions from \textit{SIG-Guided MCTS} via \textit{Trajectory-based SFT}.

\noindent\textbf{SIG-Guided MCTS Sampling.}
To construct optimal interaction trajectories, we apply MCTS~\cite{coulom2006MCTS} guided by SIG, simulating a dialogue tree rooted at the initial partial information question $Q_p$. Each intermediate node $n_t = (q_t, r_t)$ represents a follow-up question $q_t$ and its corresponding response $r_t$. Each leaf node represents a final answer $A'$. Node expansion is governed by a system prompt that instructs the model to ask or answer based on current information sufficiency. The SIG reward (Eq~\ref{eq:sig_reward}) guides exploration for better paths by quantifying the clinical value of each questioning node.
The MCTS proceeds through the following steps (see Appendix~\ref{appendix:mcts_sampling} for details, pseudo-codes and complexity analysis):
\begin{itemize}[leftmargin=*,noitemsep,topsep=2pt]
    \item \textbf{Selection.} Select a path via Upper Confidence Bound for Trees (UCT)~\cite{kocsis2006UCT}.
    \item \textbf{Expansion.} From node $n_{t-1}$, the model either generates a follow-up question $q_{t}$ and receives $r_{t}$ to expand node $n_{t}$, or predicts final answer $A'$. 
    \item \textbf{Simulation.} Interaction continues until termination or a depth limit, with accumulated $SIG(q_{t})$ rewards and a final correctness reward for $A'$.
    \item\textbf{Backpropagation.} The total reward of the trajectory is propagated to update all visited nodes.
\end{itemize}
The overall reward for a complete interaction trajectory $\tau = \{Q_p, (q_1,r_1),\dots,(q_T,r_T), A'\}$ is:
\begin{equation}
\label{eq:trajectory_reward}
\small
R(\tau) =  \alpha \cdot \mathbf{1}(A' = A^*) + \beta  \sum_{t=1}^{T} \text{SIG}(q_t),
\end{equation}
where $\alpha$ and $\beta$ are coefficients controlling outcome and process reward. The search process is conducted on patient cases in the training data. We retain the answer-correct trajectory with the highest $R(\tau)$ for each patient case.

\noindent\textbf{Trajectory-Based SFT.}
We fine-tune the LLM on selected high-reward trajectories to imitate the optimal information-seeking behavior, learning when to ask, what to ask, and when to answer. We supervise only the model-generated tokens, i.e., the follow-up questions $\{q_1,\dots,q_T\}$ and the final answer $A'$, while masking out patient responses and prompts during loss computation and gradient propagation. The loss function is:
\begin{equation}
\small
\begin{aligned}
\mathcal{L}_{\text{SFT}} &= \mathbb{E}_{\tau \sim \mathcal{D}_{\text{SFT}}} \left[
    \sum_{t=1}^{T} \mathcal{L}_{\text{question}}^{(t)} + \mathcal{L}_{\text{answer}}
\right],\\
\mathcal{L}_{\text{question}}^{(t)} &= -\log P_\theta(q_t \mid Q_p, \{(q_{i}, r_{i})\}_{i=1}^{t-1}) \\
\mathcal{L}_{\text{answer}} &= -\log P_\theta(A' \mid Q_p, \{(q_{i}, r_{i})\}_{i=1}^{T})
\end{aligned}
\end{equation}

\subsection{SIG-Augmented Policy Optimization}

This stage further enhances the model's proactive information-seeking ability via RL by extending GRPO with SIG reward. A novel \textit{SIG-Guided Reward Distribution Mechanism} decomposes trajectory-level rewards into action-level signals, prioritizing clinically valuable questions for targeted, fine-grained optimization.

\noindent\textbf{SIG-Guided Reward Distribution Mechanism.}
Following GRPO, for each partial information question $Q_p$, a group of trajectories $\mathcal{G} = \{\tau_1, \dots, \tau_K\}$ is sampled from the current policy $\pi_{\theta_{old}}$. The trajectory-level reward $R(\tau_i)$ is computed via Eq~\ref{eq:trajectory_reward}, capturing both the outcome correctness and cumulative information gain.

In standard GRPO, the trajectory-level reward is used to derive the group-relative advantage by comparing the performance of trajectories within the group. This advantage is uniformly assigned to all model-generated tokens, assuming that each token, whether part of a question or the final answer, contributes equally to the outcome. While this approach captures the overall quality of the trajectory, it overlooks its internal heterogeneity: some questions may elicit more clinical information while others may be redundant or irrelevant. Consequently, such equal feedback fails to prioritize questions with higher clinical values.

To address this, we introduce the SIG-Guided Reward Distribution Mechanism, decomposing the trajectory-level reward $R(\tau)$ into action-specific rewards for each question and the final answer:
\begin{itemize}[leftmargin=*,noitemsep,topsep=2pt]
    \item Each follow-up question $q_t$ receives:
    \begin{equation}
\small
\begin{aligned}
R(q_t) = \beta \cdot \text{SIG}(q_t) &+ \lambda_q \cdot w_t \cdot \mathbf{1}(A' = A^*),\\
\text{where}\quad w_t = &\frac{\text{SIG}(q_t)}{\sum_{j=1}^{T} \text{SIG}(q_j)}.
\end{aligned}
\end{equation}
    \item The final answer $A'$ receives:
    \begin{equation}
    \small
    R(A') = \lambda_a \cdot \mathbf{1}(A' = A^*).
    \end{equation}
\end{itemize}
Here, $\lambda_q + \lambda_a = \alpha$ ensures the total correctness reward is preserved and fully distributed across actions. The normalized SIG weight $w_t$ reflects the relative contribution of each question to the final answer. This decomposition guarantees that action-level rewards add up to the trajectory reward: 
\begin{equation}
\small
\sum_{t=1}^{T} R(q_t) + R(A') = R(\tau)
\end{equation}
To provide token-level feedback, action rewards are further propagated to individual tokens. Let $\{x_1, x_2, \dots, x_N\}$ denote the token sequence of trajectory $\tau$. Each token $x_i$ inherits the reward of the action that it belongs to:
\begin{equation}
\small
\begin{aligned}
r(x_i) &=
\begin{cases}
R(q_i), & \text{if } x_i \in q_t \\
R(A'), & \text{if } x_i \in A' \\
0, & \text{otherwise}
\end{cases} \\
\end{aligned}
\label{eq:token_level_reward}
\end{equation}
This assignment ensures that actions providing more clinical utility receive proportionally higher rewards, encouraging informative questions that contribute meaningfully to the correct outcome.

\noindent\textbf{Final Optimization Objective.}
Next, we normalize token-level rewards across the group $\mathcal{G}$ to obtain group-relative token-level advantages. Let $\mathcal{R}_{\mathcal{G}} = \{ r(x_i) \mid x_i \in \tau_k, \tau_k \in \mathcal{G} \}$ be the set of all token rewards across the group. The advantage $\hat{A}(x)$ for each token $x$ is:
\begin{equation}
\small
\hat{A}(x_i) = \frac{r(x_i) - \text{mean}(\mathcal{R}_{\mathcal{G}})}{\text{std}(\mathcal{R}_{\mathcal{G}})},
\end{equation}
Finally, we apply token-level advantages to the optimization objective. Let $\hat{A}_{k,i}$ denote the advantage of token $x_i$ in trajectory $\tau_k$. The objective is:
\begin{equation}
\small
\begin{aligned}
&\mathcal{J}(\theta)
= \mathbb{E}_{Q_p \sim \mathcal{D},\; \{\tau_k\}_{k=1}^K \sim \pi_{\theta_{\text{old}}}} \\
&\Bigg[\frac{1}{K} \sum_{k=1}^{K} \frac{1}{|\tau_k|} \sum_{i=1}^{|\tau_k|}
\min \big( r_{k,i} \hat{A}_{k,i},\;
\text{clip}(r_{k,i}, 1 \pm \epsilon) \hat{A}_{k,i} \big)\Bigg] \\
\end{aligned}
\end{equation}
where the importance ratio is defined as:
\begin{equation}
\small
r_{k,i} = \frac{\pi_\theta(\tau_{k,i} \mid Q_p, \tau_{k,<i})}{\pi_{\theta_{\text{old}}}(\tau_{k,i} \mid Q_p, \tau_{k,<i})}.
\label{eq:ratio_final}
\end{equation}
Here, $\pi_\theta$ is the policy model, and $\tau_{k,<i}$ the decoding context preceding token $x_i$. This objective assigns fine-grained credit within trajectories, providing differentiated gradients at the token level.

\noindent\textbf{Model-Aware Dynamic Rewarding.}
Rather than static precomputation, the atomic fact Shapley values within our SIG reward are dynamically computed during training. By recalculating these values based on the model’s evolving prediction probabilities (Eq~\ref{eq:fact_shapley}) at each update step, the SIG reward remains strictly model-aware. This ensures the reward accurately captures the value of information relative to the model’s current state.

\section{Experiments}

\begin{table*}[ht]
\centering
\fontsize{9pt}{9pt}\selectfont
\renewcommand{\arraystretch}{1}
\resizebox{\linewidth}{!}
{
\begin{tabular}{c|l|cc|cc|cc}
\toprule
\rowcolor[gray]{0.95}
 \multicolumn{2}{c|}{\textbf{\textit{LLM Turbo}}} & \multicolumn{2}{c|}{\textbf{\textit{Qwen3-1.7B}}} & \multicolumn{2}{c|}{\textbf{\textit{LLaMA3.2-3B}}} & \multicolumn{2}{c}{\textbf{\textit{LLama3.1-8B}}} \\
\rowcolor[gray]{0.95}\textbf{Type}& \textbf{Method} & MedQA $\uparrow$ & CMB-Exam $\uparrow$ & MedQA $\uparrow$ & CMB-Exam $\uparrow$ & MedQA $\uparrow$ & CMB-Exam $\uparrow$\\
\midrule
\multirow{7}{*}{Prompt} 
& Direct & \pmstd{36.34}{1.34} & \pmstd{19.34}{0.91} & \pmstd{27.48}{1.23} & \pmstd{43.10}{1.10} & \pmstd{48.32}{1.41} & \pmstd{44.10}{1.14}  \\
& Vanilla & \pmstd{31.05}{1.30} & \pmstd{29.01}{1.02} & \pmstd{30.81}{1.26} & \pmstd{33.45}{1.07} & \pmstd{40.11}{1.32} & \pmstd{44.49}{1.10}  \\
& CoT & \pmstd{29.54}{1.23} & \pmstd{31.04}{1.05} & \pmstd{35.61}{1.33} & \pmstd{34.79}{1.13} & \pmstd{43.12}{1.41} & \pmstd{44.75}{1.13}  \\
& MCTS-BT & \pmstd{29.77}{1.29} & \pmstd{30.28}{1.05} & \pmstd{30.24}{1.32} & \pmstd{21.50}{0.95} & \pmstd{34.41}{1.35} & \pmstd{42.95}{1.15} \\
& MCTS-MV & \pmstd{33.39}{1.33} & \pmstd{37.93}{1.06} & \pmstd{30.64}{1.26} & \pmstd{30.03}{1.02} & \pmstd{42.50}{1.43} & \pmstd{50.44}{1.18} \\
& MEDIQ & \pmstd{37.13}{1.44} & \pmstd{31.13}{1.04} & \pmstd{35.73}{1.50} & \pmstd{15.97}{0.77} & \pmstd{44.78}{1.95}  & \pmstd{33.16}{1.40}   \\
& UoT & \pmstd{36.94}{1.30} & \pmstd{43.79}{0.99}  & \pmstd{36.71}{1.18} & \pmstd{40.57}{1.01}  
& \pmstd{36.68}{1.32} & \pmstd{44.04}{1.02}    \\
\midrule
\multirow{4}{*}{SFT} 
& DialogT & \pmstd{28.54}{1.23} & \pmstd{32.43}{1.07} & \pmstd{28.83}{1.35} & \pmstd{34.07}{1.08} & \pmstd{33.42}{1.36} & \pmstd{38.37}{1.10} \\
& SFT-GT & \pmstd{33.49}{1.25} & \pmstd{43.65}{1.13} & \pmstd{42.76}{1.38} & \pmstd{42.37}{1.11} & \pmstd{48.74}{1.40} & \pmstd{49.93}{1.13}\\
& SFT & \pmstd{36.85}{1.36} & \pmstd{44.30}{1.10} & \pmstd{41.83}{1.35} & \pmstd{43.60}{1.13} & \pmstd{49.39}{1.43} & \pmstd{47.15}{1.12}\\
& \textbf{\M(S\#1)} & \pmstd{37.61}{1.37} & \pmstd{45.69}{1.16} & \pmstd{43.69}{1.37} & \underline{\pmstd{45.07}{1.17}} & \pmstd{52.63}{1.44} & \underline{\pmstd{51.78}{1.15}} \\
\midrule
\multirow{3}{*}{SFT+RL}
& \M(S\#1)+DPO & \underline{\pmstd{38.06}{1.37}} & \pmstd{42.42}{1.09} & \pmstd{43.44}{1.46} & \pmstd{42.87}{1.14} & \pmstd{52.80}{1.40} & \pmstd{47.87}{1.14}  \\
& \M(S\#1)+GRPO & \pmstd{37.62}{1.39} & \underline{\pmstd{46.61}{1.22}} & \underline{\pmstd{46.32}{1.37}} & \pmstd{44.76}{1.31} & \underline{\pmstd{54.60}{1.37}} & \pmstd{51.43}{1.13}\\
& \textbf{\M(S\#1+2)} & \textbf{\pmstd{39.93}{1.43}} & \textbf{\pmstd{51.98}{1.17}} & \textbf{\pmstd{47.38}{1.54}} & \textbf{\pmstd{46.25}{1.13}} & \textbf{\pmstd{55.60}{1.30}} & \textbf{\pmstd{59.33}{1.11}} \\
\midrule
\rowcolor[gray]{0.95} 
\multicolumn{2}{c|}{\textit{*Performance Gain (\%)}} 
& \textit{+4.91} & \textit{+11.52} & \textit{+2.29} & \textit{+2.62} & \textit{+1.83} & \textit{+14.58} \\
\bottomrule
\end{tabular}
}
\caption{
Performance comparison (\%) on \textbf{MedQA} and \textbf{CMB-Exam}. \textbf{Bold} indicates the best performance, \underline{underline} the second-best. Performance gains are computed as the relative improvements over the second-best performances.}
\label{tab:comparison}
\end{table*}

\subsection{Experimental Setup}
\noindent\textbf{Datasets.} Experiments are conducted on two datasets derived from public multiple-choice medical benchmarks: \textbf{MedQA}~\cite{jin2021medqa} and \textbf{CMB}~\cite{wang2024cmb}. Each original question is decomposed into atomic facts $\mathcal{F}$ and an atomic question $Q$ that excludes factual information. The model input is a partial information question $Q_p$, consisting of $Q$ and a subset of $\mathcal{F}$: 50\% facts for CMB and the chief complaint for MedQA (following MEDIQ~\cite{li2024mediq}). Dataset construction and statistics can be found in Appendix~\ref{appendix:dataset}.

\noindent\textbf{Baselines.} We compare with diverse baselines:
\begin{itemize}[leftmargin=*,noitemsep,topsep=2pt]
    \item\textbf{Prompt-based.} \textbf{Direct} generates answers without interaction. \textbf{Vanilla} uses a system prompt to encourage questioning. \textbf{COT} adds ``Let's think step by step'' to promote reasoning. \textbf{MCTS-BT} uses MCTS with self-evaluated reward for inference-time scaling and selects the best trajectory, while \textbf{MCTS-MV} adopts majority voting over sampled trajectories. \textbf{MEDIQ}~\cite{li2024mediq} implements an abstention module. \textbf{UoT}~\cite{hu2024UoT} selects questions by maximizing expected entropy reduction.
    \item \textbf{SFT-based.} \textbf{DialogT}~\cite{liu2025dialogueT} reformulates QAs as dialogues for fine-tuning. \textbf{SFT-GT} uses gold answers for supervision. \textbf{SFT} samples correct trajectories without SIG-guided MCTS.
    \item \textbf{SFT+RL.} RL is conducted upon \M~(S\#1) initialization for fair comparisons. \textbf{DPO} contrasts correct and incorrect trajectories sampled from the model. \textbf{GRPO} uses trajectory-level correctness reward.
\end{itemize}

\noindent \textbf{Evaluation Metrics.} 
Exact Match (EM) accuracy is reported, where options (e.g., "A", "CD") extracted from the model's final answer must be identical to the ground-truth set.
\begin{equation*}
\small
Accuracy = \frac{\text{Correct Predictions}}{\text{Total Questions}} \times 100
\end{equation*}

\noindent\textbf{Implementations.}
Experiments are conducted on \textit{Instruct} LLMs: \textit{LLaMA3.1-8B}, \textit{LLaMA3.2-3B}~\cite{dubey2024llama3}, and \textit{Qwen3-1.7B}~\cite{yang2025qwen3}, with patient simulator by \textit{Qwen2.5-72B} and SFT data sampled from \textsc{Deepseek-R1}~\cite{guo2025deepseek}. Training-based methods are trained on benchmark-specific training sets. Prompt-based methods are directly tested. Implementation details are in Appendix~\ref{appendix:implementation}.

\begin{table}[t]
\centering
\fontsize{9pt}{9pt}\selectfont
\renewcommand{\arraystretch}{1}
\begin{tabular}{c|cc}
\toprule
\rowcolor[gray]{0.95}\textbf{Ablation} & \textbf{MedQA $\uparrow$} & \textbf{CMB (OOD) $\uparrow$} \\
\midrule
Vanilla & 40.11 & 44.49 \\
\midrule
w/o Stage 1 & 35.59 & 43.78 \\
w/o Stage 2 & 53.26 & 41.44 \\
\midrule
w/o SIG & 54.42 & 40.83 \\
w/o Shapley & 54.55 & 42.73 \\
w/o Distribution & \underline{54.60} & \underline{45.00} \\
\midrule
\textbf{\M} & \textbf{55.60} & \textbf{45.48}\\ 
\bottomrule
\end{tabular}
\caption{Ablation studies of \M.}
\label{tab:main_ablations}
\end{table}

\subsection{Main Results}
Experiment results are shown in Table~\ref{tab:comparison}. We summarize the key findings below:

\noindent \textbf{Training for proactive questioning is essential.} Direct answering without acquiring information yields poor accuracy (e.g., 19.34\% on CMB, \textit{Qwen3-1.7B}), demonstrating the necessity to address the reactive paradigm of LLMs under information insufficiency. Prompt-based methods, including advanced frameworks like UoT and MEDIQ, fail to consistently outperform direct answering (e.g., UoT on LLaMA3.2-3B MedQA). These results highlight the limitations of prompting and the importance of targeted training.

\noindent \textbf{\M~significantly outperforms existing methods and enhances LLMs' proactive information-seeking ability.} \M~consistently achieves the highest accuracy, surpassing baselines by an average relative improvement of \textbf{6.29\%} over second-best results, and a striking \textbf{54.45\%} gain over direct answering. This demonstrates that \M~effectively shifts LLMs from passively reacting to proactively acquiring information, which supports its potential for practical clinical consultations. 

\noindent \textbf{\M(S\#1) provides high-quality supervision via SIG-guided MCTS.} Among SFT methods, \M(S\#1) leverages SIG-guided MCTS to sample clinically valuable interactions and achieves the best performance. In contrast, DialogT constructs general multi-turn dialogues that fail to target information gaps. Standard SFT, which retains answer-correct samples without assessing question quality, also underperforms. To ensure the performance gain stems from improved questioning rather than answer memorization, we also fine-tune the model using ground-truth answers (SFT-GT) instead of sampled trajectories. Despite leveraging more data, SFT-GT still underperforms~\M, confirming that \M(S\#1) offers higher-quality supervision and a stronger initialization for proactive ability.

\noindent \textbf{\M(S\#2) further boosts proactive ability.}  Starting from the same~\M(S\#1) initialization, our SIG-Augmented Policy Optimization consistently outperforms other RL methods. While DPO and GRPO occasionally fail to improve the SFT-model, Stage 2 offers stable and significant gains, underscoring the benefit of our tailored SIG reward and reward distribution in optimizing the model’s information-seeking strategy.

Supplementary experiments, including hyperparameter and Shapley analyses are in Appendix~\ref{appendix:supplementary_experiments}.
Runtime analysis is provided in Appendix~\ref{sec:framework_complexity}.

\subsection{Ablation Studies}
We systematically ablate key components of \M~to validate their effectiveness. Experiments are conducted on MedQA-trained LLaMA3.1-8B, and evaluated on both in-domain and OOD settings (Table~\ref{tab:main_ablations}). The OOD setting corresponds to a train-test distribution shift, where models trained on MedQA are evaluated on CMB-Exam, referred to as CMB (OOD).

\textbf{Both stages are essential and complementary.}
Removing either SIG-guided Model Initialization or SIG-Augmented Policy Optimization leads to substantial performance drops. Notably, removing Stage~1 causes the largest in-domain drop, highlighting that a good initialization is crucial for avoiding poor RL convergence. Removing Stage~2 severely hurts OOD performance, confirming the importance of SIG-based policy optimization for cross-distribution generalization.

\textbf{Each component in the reward design contributes to model performance.}
Removing the process reward SIG reduces both in-domain performance and OOD generalization. Removing the Shapley weighting or the reward distribution mechanism also leads to consistent drops, confirming the importance of modeling question utility and allocating fine-grained reward accordingly.

\section{Analysis}

\subsection{Out-of-Domain Generalization}

\begin{table*}[t]
\centering
\small
\resizebox{\linewidth}{!}
{
\begin{tabular}{l|cc|cc|ccc}
\toprule
\rowcolor[gray]{0.95}
 & \multicolumn{2}{c|}{\textbf{MedQA} (Acc. $\uparrow$)} & \multicolumn{2}{c|}{\textbf{CMB-Clin} (Gen. $\uparrow$)} & \multicolumn{3}{c}{\textbf{Abg-CoQA} (Action Level Acc. $\uparrow$)} \\
\rowcolor[gray]{0.95} \multirow{-2}{*}{\textbf{Method}} & \textbf{MCQ} & \textbf{Open-Ended} & \textbf{BLEU-4} & \textbf{ROUGE-L} & \textbf{Ambiguous} & \textbf{Non-Ambiguous} & \textbf{Macro-Acc} \\ \midrule
Vanilla & 40.11 & 20.35 & 57.25 & 19.41 & 60.16 & \textbf{48.11} & \underline{54.15} \\
\M(S\#1)+DPO & 49.88 & 21.29 & \underline{57.79} & 20.45 & \textbf{93.44} & 9.11 & 51.28 \\
\M(S\#1)+GRPO & \underline{51.29} & \underline{21.92} & 50.34 & \underline{31.00} & 77.24 & 30.93 & 54.08 \\ \midrule
\textbf{\M(S\#1+2)} & \textbf{57.50} & \textbf{25.37} & \textbf{60.90} & \textbf{36.18} & \underline{79.51} & \underline{34.30} & \textbf{56.91} \\ \bottomrule
\end{tabular}
}
\small
\caption{Results of OOD evaluations. CMB-trained LLaMA-8B is evaluated in cross-task, cross-domain settings. \textbf{Bold} indicates the best performance, and \underline{underline} indicates the second-best performance.}
\label{tab:ood_test}
\end{table*}

\begin{table}[t]
\centering
\small
\begin{tabular}{lcc}
\toprule
\rowcolor[gray]{0.95}
\textbf{Patient Simulator Setting} & \textbf{MedQA} & \textbf{CMB (OOD)} \\
\midrule
\multicolumn{3}{l}{\emph{Backbone Replacement}} \\
\quad \textsc{Qwen2.5-72B} (default) &  55.60 & \textbf{45.48} \\
\quad \textsc{Qwen3-32B}            & 56.23 & 44.35\\
\quad \textsc{DeepSeek-V3.2}        & 56.77 & 44.83\\
\quad \textsc{GPT-5}               & 57.23 & 44.14\\
\midrule
\multicolumn{3}{l}{\emph{Enhanced Simulation Mechanisms (Qwen3-32B)}} \\
\quad + Fact-checker               & \textbf{59.07} & \underline{45.18} \\
\quad + MBTI persona   & \underline{58.17} & 44.32 \\
\bottomrule
\end{tabular}
\caption{Sensitivity analysis on patient simulators.}
\label{tab:patient_simulator_sensitivity}
\end{table}

To verify that \M~ enhances intrinsic proactivity rather than task-specific overfitting, we conduct comprehensive OOD evaluations on CMB-trained LLaMA3.1-8B across \textit{tasks} and \textit{domains}:
\begin{itemize}[leftmargin=*,noitemsep,topsep=2pt]
\item\textbf{Multiple-Choice Evaluation.} We evaluate zero-shot performance on the unseen MedQA. 

\item\textbf{Open-Ended Medical QA.}
We examine whether the learned proactive behavior generalizes to open-ended tasks.
\textbf{(1) MedQA without options.}
We remove answer options and convert MedQA into a free-form generation task.
\textbf{(2) Clinical Reasoning}
CMB-Clin~\cite{wang2024cmb} consists of EHR-based complex clinical analysis (e.g., differential diagnosis, treatment planning), measured by BLEU-4 and ROUGE against reference expert answers.

\item \textbf{General-Domain QA.}
We test whether \M~ generalizes beyond healthcare on general-domain Abg-CoQA~\cite{guo2021abg-coqa}, assessing action-level accuracy in distinguishing between ambiguous (requiring clarification) and non-ambiguous queries.

\item \textbf{Heterogeneous Clinical Data.}
We include OOD tests on MIMIC-IV in Appendix~\ref{appendix:mimic4}.
\end{itemize}

\noindent\textbf{Results.}
As shown in Table~\ref{tab:ood_test}, \M~consistently outperforms all baselines across OOD medical tasks (57.50\% MCQ, 25.37\% Open-Ended for MedQA, 36.18 ROUGE for CMB-Clin), demonstrating robust reasoning that transfers to complex realistic clinical scenarios. On general-domain Abg-CoQA, \M~effectively handles ambiguity, achieving the highest Macro-Acc. Its occasional over-questioning in non-ambiguous cases reflects a "cautious-proactive" strategy, prioritizing information gathering over premature answering, a desirable property for high-stakes domains. Overall, these results confirm that \M~effectively shifts the model paradigm from reactive to proactive.

\subsection{Varying Initial Information Amount}
\begin{figure}[t]
    \centering    \includegraphics[width=\linewidth]{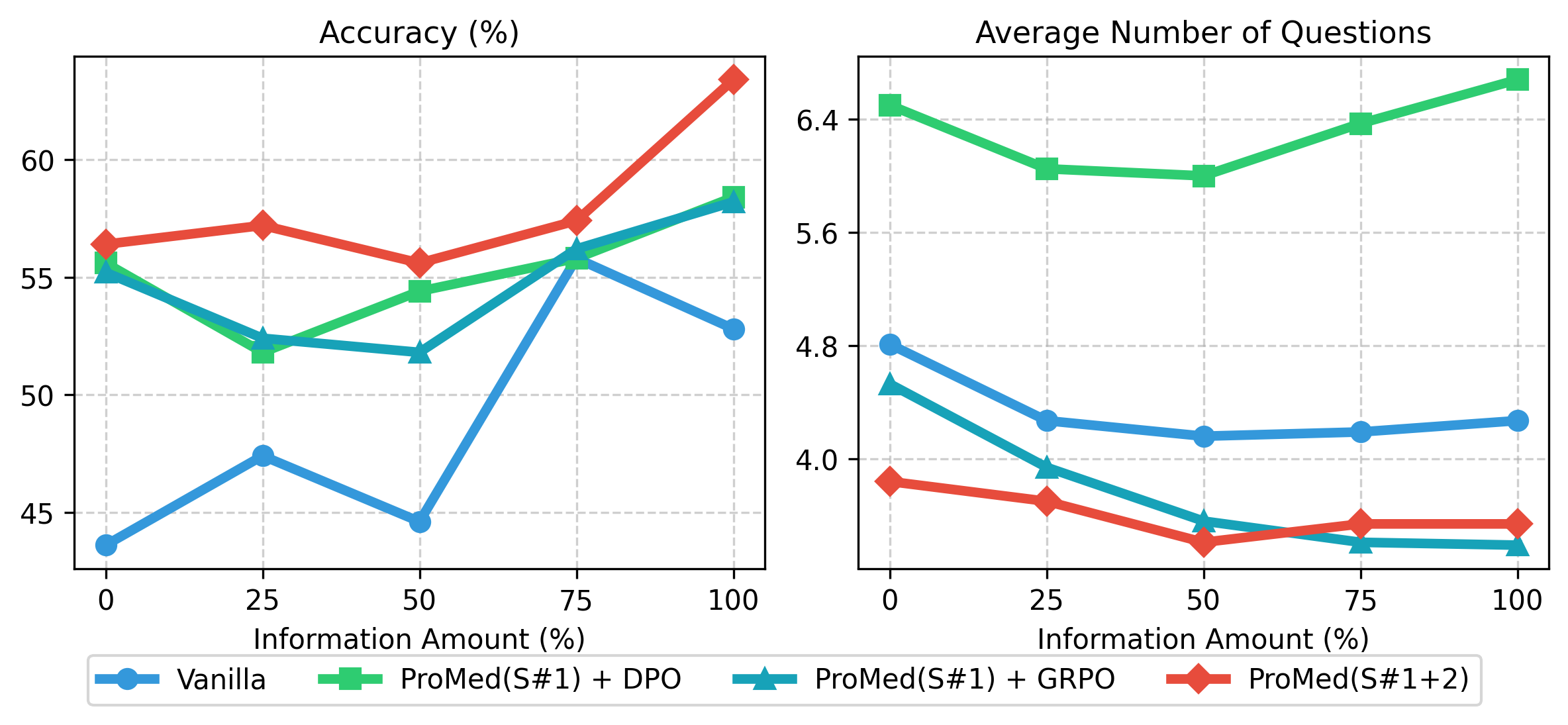}
    \caption{
    Model performance on CMB and question efficiency under varying initial information amounts.}
    \label{fig:vary_info}
\end{figure}
To investigate models' proactive behavior across different levels of information scarcity, we vary the proportion of initial facts from 0\% to 100\% on 500 sampled CMB test instances and evaluate CMB-tuned LLaMA3.1-8B (Figure~\ref{fig:vary_info}).

\textbf{\M~consistently achieves superior accuracy and questioning efficiency across all information levels}, demonstrating robust generalization regardless of how much prior context is provided. In contrast, the untrained Vanilla struggles under low-information settings (0\%–50\%), indicating its limited questioning ability to compensate for missing evidence, while baselines initialized with \M~Stage 1 SFT deliver better results. DPO relies on relatively more questions while \M~Stage 2 exhibits targeted and information-efficient behavior, reaching higher accuracy with fewer questions. These results highlight the robustness and flexibility of \M, which successfully trains LLMs to dynamically adjust questioning strategies to gather essential information.

\subsection{Robustness to Patient Simulator Variations}
To assess potential bias from LLM-simulated patients and ensure robustness to simulator choice, we evaluate MedQA-trained LLaMA3.1-8B under diverse simulator configurations, including:
\begin{itemize}[leftmargin=*,noitemsep,topsep=2pt]
\item varying the simulator’s LLM backbone;
\item introducing enhanced mechanisms, including a fact-checker to mitigate hallucinations and MBTI personas for diverse role-playing behaviors.
\end{itemize}
Detailed configurations and simulator quality analyses are reported in Appendix~\ref{appendix:patient_simulator}.
As shown in Table \ref{tab:patient_simulator_sensitivity}, performance remains remarkably stable across all simulator variants on both in-domain MedQA and OOD CMB. Adding a Fact-checker slightly improves test time performance, suggesting that higher-quality feedback during interaction further benefits the model. The results confirm that \M's effectiveness does not depend on a specific patient simulator design during training.

\subsection{Intermediate Question Quality}
\begin{figure}[t]
    \centering
    \includegraphics[width=\linewidth]{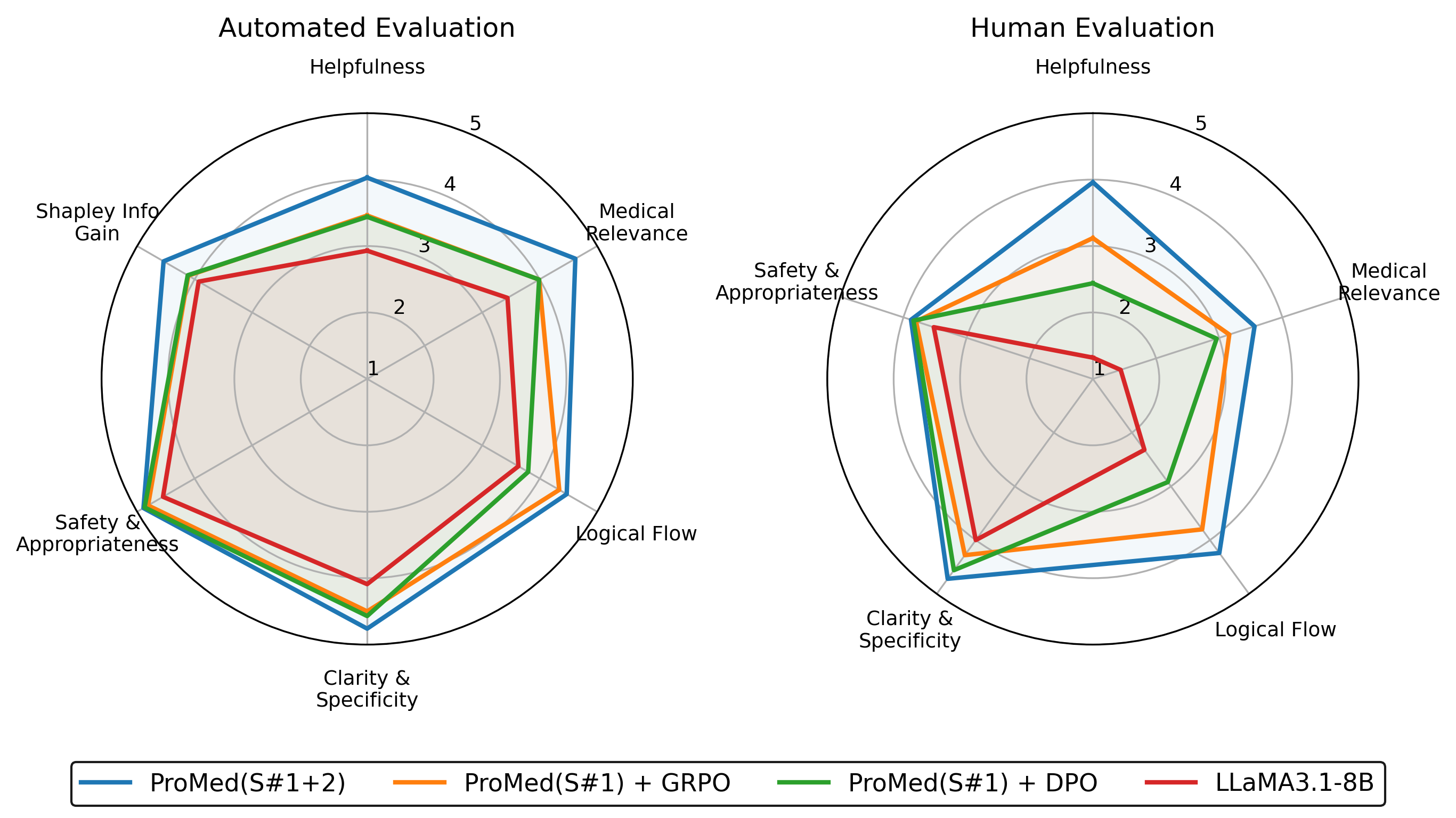}
    \caption{Question quality by both automated evaluations (left) and human clinicians (right).}
    \label{fig:question_quality_radar}
\vspace{-0.2cm}
\end{figure}

Beyond final answer accuracy, we analyze the quality of model-generated questions using both LLM-based evaluators and human clinicians. Each trajectory is evaluated along five clinically motivated dimensions: \textbf{Helpfulness}, \textbf{Medical Relevance}, \textbf{Logical Flow}, \textbf{Clarity and Specificity}, and \textbf{Safety and Appropriateness} (detailed evaluation protocols in Appendix~\ref{appendix:question_evaluation}). We additionally report the accumulated SIG reward over each trajectory as a quantitative measure of question values.

As shown in Figure~\ref{fig:question_quality_radar}, untrained models produce low-utility questions, especially in Helpfulness and Medical Relevance, reflecting generic, weakly-grounded questions. In contrast, \M-initialized and RL-optimized models show improved question quality across all dimensions. Notably, \M(S\#1+2) achieves the best overall performance, indicating a stronger ability to identify and query missing critical information. Improvements in Logical Flow suggest that the model better conditions its questions on prior evidence, thereby reducing redundancy and enhancing coherence. Although clinicians apply stricter standards, the trends remain consistent. Overall, \M~substantially improves the \emph{clinical value} and \emph{strategic quality} of proactive questions, enabling more effective and reliable consultations.

Qualitative examples and analyses of representative interaction trajectories are in Appendix~\ref{appendix:case_study}.

\section{Conclusions and Future Works}
We present \M, a novel RL framework that shifts medical LLMs from reactive to proactive for interactive medical consultation. By introducing SIG reward, \M~quantifies question utility while accounting for interactions among clinical information. Building on this, a two-stage SIG-augmented RL pipeline enables stable initialization and fine-grained optimization via a reward distribution mechanism. Experiments show that \M~outperforms existing methods and generalizes well to OOD settings. Future work will extend this paradigm to long-term reasoning and broader interactive decision-making tasks.

\section*{Limitations}

While \M~framework offers a promising step toward proactive medical LLMs, enabling them to actively seek information in clinical settings, several limitations remain. First, training and evaluation are conducted in simulated dialogue environments, which may not fully capture the complexity and variability of real-world patient interactions. Second, although our experiments primarily focus on two multiple-choice clinical benchmarks covering diagnosis, medication, and test recommendation, we also validate the effectiveness of our approach on several open-ended medical questions and general-domain tasks. Nevertheless, further evaluation on more diverse, complex, and real-world clinical scenarios, such as free-form treatment planning or multi-patient longitudinal cases, is necessary. Third, due to computational resource constraints, we use models up to 8B parameters; while representative, these models may not fully reveal the performance ceiling achievable with larger-scale LLMs. Finally, our current implementation operates on text-based medical facts. Incorporating multimodal clinical data, such as time series, imaging, and data from multiple comprehensive datasets, remains an important direction toward building more general and broadly applicable proactive LLMs.

\section*{Ethical considerations}
This study develops a reinforcement learning framework guided by Shapley Information Gain to enhance the proactive ability of LLMs for interactive medical consultations. All experiments are conducted on public, de-identified datasets (MedQA, CMB, CMB-Clin, and Abg-CoQA) that do not contain personally identifiable information. Human evaluations are conducted under ethical approvals. While the trained model demonstrates improved proactive questioning under partial information, it is intended purely for research purposes and is not deployed in real-world clinical scenarios. The model is not designed to replace medical professionals, and any future application would require rigorous clinical validation, safety testing, and adherence to medical regulatory standards. We note that some datasets may reflect population-specific characteristics: for example, the Chinese datasets primarily represent East Asian populations, whereas the English datasets mainly reflect Western populations. Nevertheless, our proposed framework is model-agnostic and task-agnostic, and the underlying methodology is generalizable across populations and clinical contexts. We believe this work supports the responsible advancement of medical AI by addressing the risks of hallucination and unreliable responses that may arise in reactive medical LLMs when operating under incomplete patient information.

\section*{Acknowledgments}
This work is supported by the National Natural Science Foundation of China (No.62576013, No.U23A20468), Prevention and Control of Emerging and Major Infectious Diseases-National Science and Technology Major Project (2025ZD01906000, 2025ZD01906004), the National Natural Science Foundation of China (62402017), Peking University Clinical Medicine Plus X Pilot Program-Key Technologies Project (2024YXXLHGG007), and “TengYun” Clinical Research Program (TY2025015). 

Liantao Ma is supported by the Beijing Traditional Chinese Medicine Science and Technology Development Fund (BJZYZD-2025-13), the Young Elite Scientists Sponsorship Program of the Beijing High Innovation Plan (20250628).




\bibliography{custom}

@article{johnson2023mimic,
  title={MIMIC-IV, a freely accessible electronic health record dataset},
  author={Johnson, Alistair EW and Bulgarelli, Lucas and Shen, Lu and Gayles, Alvin and Shammout, Ayad and Horng, Steven and Pollard, Tom J and Hao, Sicheng and Moody, Benjamin and Gow, Brian and others},
  journal={Scientific data},
  volume={10},
  number={1},
  pages={1},
  year={2023},
  publisher={Nature Publishing Group UK London}
}

@inproceedings{lin2004rouge,
  title={Rouge: A package for automatic evaluation of summaries},
  author={Lin, Chin-Yew},
  booktitle={Text summarization branches out},
  pages={74--81},
  year={2004}
}

@inproceedings{papineni2002bleu,
  title={Bleu: a method for automatic evaluation of machine translation},
  author={Papineni, Kishore and Roukos, Salim and Ward, Todd and Zhu, Wei-Jing},
  booktitle={Proceedings of the 40th annual meeting of the Association for Computational Linguistics},
  pages={311--318},
  year={2002}
}

@article{eastmond1979nail1,
  title={The nail dystrophy of psoriatic arthritis.},
  author={Eastmond, CJ and Wright, V},
  journal={Annals of the Rheumatic Diseases},
  volume={38},
  number={3},
  pages={226--228},
  year={1979},
  publisher={BMJ Publishing Group Ltd}
}

@article{jha2019nail2,
  title={Nail pitting: a key clinical sign of psoriatic arthritis},
  author={Jha, Saket and Kumar, Rajiv Ranjan and Dhooria, Aadhaar and Sharma, Aman},
  journal={Rheumatology},
  volume={58},
  number={12},
  pages={2250--2250},
  year={2019},
  publisher={Oxford University Press}
}

@misc{deepseekai2025deepseekv32,
      title={DeepSeek-V3.2: Pushing the Frontier of Open Large Language Models}, 
      author={DeepSeek-AI},
      year={2025},
}

@inproceedings{kwon2023efficient,
  title={Efficient Memory Management for Large Language Model Serving with PagedAttention},
  author={Woosuk Kwon and Zhuohan Li and Siyuan Zhuang and Ying Sheng and Lianmin Zheng and Cody Hao Yu and Joseph E. Gonzalez and Hao Zhang and Ion Stoica},
  booktitle={Proceedings of the ACM SIGOPS 29th Symposium on Operating Systems Principles},
  year={2023}
}

@inproceedings{guo2021abg-coqa,
  title={Abg-coqa: Clarifying ambiguity in conversational question answering},
  author={Guo, Meiqi and Zhang, Mingda and Reddy, Siva and Alikhani, Malihe},
  booktitle={3rd Conference on Automated Knowledge Base Construction},
  year={2021}
}

@article{snyder2018acuteappendicitis,
  title={Acute appendicitis: efficient diagnosis and management},
  author={Snyder, Matthew J and Guthrie, Marjorie and Cagle, Stephen},
  journal={American family physician},
  volume={98},
  number={1},
  pages={25--33},
  year={2018}
}

@article{balafar2024chestCTcomparison,
  title={Comparison of the diagnostic value of ultrasound with chest CT scan in patients with unspecified pulmonary pneumonia in the emergency department},
  author={Balafar, Moloud and Pouraghaei, Mahboub and Ranjkesh, Mahnaz and Dehghan, Mahshid and Delkhorrami, Ali and Shams Vahdati, Samad},
  journal={Journal of Emergency Practice and Trauma},
  volume={9},
  number={2},
  pages={87--91},
  year={2024},
  publisher={Kerman University of Medical Sciences}
}

@article{OpenAI2023GPT4TR,
  title={GPT-4 Technical Report},
  author={OpenAI},
  journal={ArXiv},
  year={2023},
  volume={abs/2303.08774}
}

@article{dubey2024llama3,
  title={The llama 3 herd of models},
  author={Dubey, Abhimanyu and Jauhri, Abhinav and Pandey, Abhinav and Kadian, Abhishek and Al-Dahle, Ahmad and Letman, Aiesha and Mathur, Akhil and Schelten, Alan and Yang, Amy and Fan, Angela and others},
  journal={arXiv e-prints},
  pages={arXiv--2407},
  year={2024}
}

@article{singhal2025medPalm,
  title={Toward expert-level medical question answering with large language models},
  author={Singhal, Karan and Tu, Tao and Gottweis, Juraj and Sayres, Rory and Wulczyn, Ellery and Amin, Mohamed and Hou, Le and Clark, Kevin and Pfohl, Stephen R and Cole-Lewis, Heather and others},
  journal={Nature Medicine},
  volume={31},
  number={3},
  pages={943--950},
  year={2025},
  publisher={Nature Publishing Group US New York}
}

@article{wu2024pmc-llama,
  title={PMC-LLaMA: toward building open-source language models for medicine},
  author={Wu, Chaoyi and Lin, Weixiong and Zhang, Xiaoman and Zhang, Ya and Xie, Weidi and Wang, Yanfeng},
  journal={Journal of the American Medical Informatics Association},
  volume={31},
  number={9},
  pages={1833--1843},
  year={2024},
  publisher={Oxford Academic}
}

@article{zhang2023huatuogpt,
  title={Huatuogpt, towards taming language model to be a doctor},
  author={Zhang, Hongbo and Chen, Junying and Jiang, Feng and Yu, Fei and Chen, Zhihong and Li, Jianquan and Chen, Guiming and Wu, Xiangbo and Zhang, Zhiyi and Xiao, Qingying and others},
  journal={arXiv preprint arXiv:2305.15075},
  year={2023}
}

@article{ding20243ds,
  title={3DS: Decomposed Difficulty Data Selection's Case Study on LLM Medical Domain Adaptation},
  author={Ding, Hongxin and Fang, Yue and Zhu, Runchuan and Jiang, Xinke and Zhang, Jinyang and Xu, Yongxin and Chu, Xu and Zhao, Junfeng and Wang, Yasha},
  journal={arXiv preprint arXiv:2410.10901},
  year={2024}
}

@inproceedings{jiang-etal-2025-hykge,
    title = "{H}y{KGE}: A Hypothesis Knowledge Graph Enhanced {RAG} Framework for Accurate and Reliable Medical {LLM}s Responses",
    author = "Jiang, Xinke  and Zhang, Ruizhe  and Xu, Yongxin  and Qiu, Rihong  and Fang, Yue  and Wang, Zhiyuan  and
      Tang, Jinyi  and Ding, Hongxin  and Chu, Xu  and
      Zhao, Junfeng  and Wang, Yasha",
    editor = "Che, Wanxiang  and Nabende, Joyce  and
      Shutova, Ekaterina  and Pilehvar, Mohammad Taher",
    booktitle = "Proceedings of the 63rd Annual Meeting of the Association for Computational Linguistics (Volume 1: Long Papers)",
    month = jul,
    year = "2025",
    address = "Vienna, Austria",
    publisher = "Association for Computational Linguistics",
    url = "https://aclanthology.org/2025.acl-long.580/",
    pages = "11836--11856",
    ISBN = "979-8-89176-251-0",
}

@article{mcduff2025diagnosis,
  title={Towards accurate differential diagnosis with large language models},
  author={McDuff, Daniel and Schaekermann, Mike and Tu, Tao and Palepu, Anil and Wang, Amy and Garrison, Jake and Singhal, Karan and Sharma, Yash and Azizi, Shekoofeh and Kulkarni, Kavita and others},
  journal={Nature},
  pages={1--7},
  year={2025},
  publisher={Nature Publishing Group UK London}
}

@inproceedings{xu2025dearllm,
  title={DearLLM: Enhancing Personalized Healthcare via Large Language Models-Deduced Feature Correlations},
  author={Xu, Yongxin and Jiang, Xinke and Chu, Xu and Qiu, Rihong and Feng, Yujie and Ding, Hongxin and Zhao, Junfeng and Wang, Yasha and Xie, Bing},
  booktitle={Proceedings of the AAAI Conference on Artificial Intelligence},
  volume={39},
  number={1},
  pages={941--949},
  year={2025}
}

@article{li2024mediq,
  title={Mediq: Question-asking llms and a benchmark for reliable interactive clinical reasoning},
  author={Li, Stella and Balachandran, Vidhisha and Feng, Shangbin and Ilgen, Jonathan and Pierson, Emma and Koh, Pang Wei W and Tsvetkov, Yulia},
  journal={Advances in Neural Information Processing Systems},
  volume={37},
  pages={28858--28888},
  year={2024}
}

@article{hu2024UoT,
  title={Uncertainty of thoughts: Uncertainty-aware planning enhances information seeking in llms},
  author={Hu, Zhiyuan and Liu, Chumin and Feng, Xidong and Zhao, Yilun and Ng, See-Kiong and Luu, Anh Tuan and He, Junxian and Koh, Pang Wei W and Hooi, Bryan},
  journal={Advances in Neural Information Processing Systems},
  volume={37},
  pages={24181--24215},
  year={2024}
}

@article{liu2025dialogueT,
  title={Dialogue is Better Than Monologue: Instructing Medical LLMs via Strategical Conversations},
  author={Liu, Zijie and Zhao, Xinyu and Peng, Jie and Zhu, Zhuangdi and Chen, Qingyu and Hu, Xia and Chen, Tianlong},
  journal={arXiv preprint arXiv:2501.17860},
  year={2025}
}

@article{liao2023automatic_consultation,
  title={An automatic evaluation framework for multi-turn medical consultations capabilities of large language models},
  author={Liao, Yusheng and Meng, Yutong and Liu, Hongcheng and Wang, Yanfeng and Wang, Yu},
  journal={arXiv preprint arXiv:2309.02077},
  year={2023}
}

@article{schulman2017ppo,
  title={Proximal policy optimization algorithms},
  author={Schulman, John and Wolski, Filip and Dhariwal, Prafulla and Radford, Alec and Klimov, Oleg},
  journal={arXiv preprint arXiv:1707.06347},
  year={2017}
}

@article{rafailov2023dpo,
  title={Direct preference optimization: Your language model is secretly a reward model},
  author={Rafailov, Rafael and Sharma, Archit and Mitchell, Eric and Manning, Christopher D and Ermon, Stefano and Finn, Chelsea},
  journal={Advances in neural information processing systems},
  volume={36},
  pages={53728--53741},
  year={2023}
}

@article{shao2024deepseekmath-grpo,
  title={Deepseekmath: Pushing the limits of mathematical reasoning in open language models},
  author={Shao, Zhihong and Wang, Peiyi and Zhu, Qihao and Xu, Runxin and Song, Junxiao and Bi, Xiao and Zhang, Haowei and Zhang, Mingchuan and Li, YK and others},
  journal={arXiv preprint arXiv:2402.03300},
  year={2024}
}

@article{yu2025dapo,
  title={Dapo: An open-source llm reinforcement learning system at scale},
  author={Yu, Qiying and Zhang, Zheng and Zhu, Ruofei and Yuan, Yufeng and Zuo, Xiaochen and Yue, Yu and Dai, Weinan and Fan, Tiantian and Liu, Gaohong and Liu, Lingjun and others},
  journal={arXiv preprint arXiv:2503.14476},
  year={2025}
}

@article{kaelbling1996reinforcement,
  title={Reinforcement learning: A survey},
  author={Kaelbling, Leslie Pack and Littman, Michael L and Moore, Andrew W},
  journal={Journal of artificial intelligence research},
  volume={4},
  pages={237--285},
  year={1996}
}

@article{sutton1999reinforcement,
  title={Reinforcement learning},
  author={Sutton, Richard S and Barto, Andrew G and others},
  journal={Journal of Cognitive Neuroscience},
  volume={11},
  number={1},
  pages={126--134},
  year={1999}
}

@article{kaufmann2024survey-RLHF,
  title={A survey of reinforcement learning from human feedback},
  author={Kaufmann, Timo and Weng, Paul and Bengs, Viktor and H{\"u}llermeier, Eyke},
  year={2024}
}

@article{wang2025healthq,
  title={Healthq: Unveiling questioning capabilities of llm chains in healthcare conversations},
  author={Wang, Ziyu and Li, Hao and Huang, Di and Kim, Hye-Sung and Shin, Chae-Won and Rahmani, Amir M},
  journal={Smart Health},
  pages={100570},
  year={2025},
  publisher={Elsevier}
}

@article{lee2025good-utility-estimation,
  title={What is a good question? utility estimation with llm-based simulations},
  author={Lee, Dong-Ho and Cho, Hyundong and May, Jonathan and Pujara, Jay},
  journal={arXiv preprint arXiv:2502.17383},
  year={2025}
}

@inproceedings{liu2025MMD-eval,
  title={Interactive evaluation for medical llms via task-oriented dialogue system},
  author={Liu, Ruoyu and Xue, Kui and Zhang, Xiaofan and Zhang, Shaoting},
  booktitle={Proceedings of the 31st International Conference on Computational Linguistics},
  pages={4871--4896},
  year={2025}
}

@article{winter2002shapley,
  title={The shapley value},
  author={Winter, Eyal},
  journal={Handbook of game theory with economic applications},
  volume={3},
  pages={2025--2054},
  year={2002},
  publisher={Elsevier}
}

@inproceedings{mazzaccara2024EIG,
  title={Learning to Ask Informative Questions: Enhancing LLMs with Preference Optimization and Expected Information Gain},
  author={Mazzaccara, Davide and Testoni, Alberto and Bernardi, Raffaella},
  booktitle={Findings of the Association for Computational Linguistics: EMNLP 2024},
  pages={5064--5074},
  year={2024}
}

@article{zhu2025ask-patient,
  title={Ask patients with patience: Enabling llms for human-centric medical dialogue with grounded reasoning},
  author={Zhu, Jiayuan and Wu, Junde},
  journal={arXiv preprint arXiv:2502.07143},
  year={2025}
}

@inproceedings{wang2024cmb,
  title={CMB: A Comprehensive Medical Benchmark in Chinese},
  author={Wang, Xidong and Chen, Guiming and Dingjie, Song and Zhiyi, Zhang and Chen, Zhihong and Xiao, Qingying and Chen, Junying and Jiang, Feng and Li, Jianquan and Wan, Xiang and others},
  booktitle={Proceedings of the 2024 Conference of the North American Chapter of the Association for Computational Linguistics: Human Language Technologies (Volume 1: Long Papers)},
  pages={6184--6205},
  year={2024}
}

@article{jin2021medqa,
  title={What disease does this patient have? a large-scale open domain question answering dataset from medical exams},
  author={Jin, Di and Pan, Eileen and Oufattole, Nassim and Weng, Wei-Hung and Fang, Hanyi and Szolovits, Peter},
  journal={Applied Sciences},
  volume={11},
  number={14},
  pages={6421},
  year={2021},
  publisher={MDPI}
}

@inproceedings{kocsis2006UCT,
  title={Bandit based monte-carlo planning},
  author={Kocsis, Levente and Szepesv{\'a}ri, Csaba},
  booktitle={European conference on machine learning},
  pages={282--293},
  year={2006},
  organization={Springer}
}

@inproceedings{coulom2006MCTS,
  title={Efficient selectivity and backup operators in Monte-Carlo tree search},
  author={Coulom, R{\'e}mi},
  booktitle={International conference on computers and games},
  pages={72--83},
  year={2006},
  organization={Springer}
}

@article{yang2025qwen3,
  title={Qwen3 technical report},
  author={Yang, An and Li, Anfeng and Yang, Baosong and Zhang, Beichen and Hui, Binyuan and Zheng, Bo and Yu, Bowen and Gao, Chang and Huang, Chengen and Lv, Chenxu and others},
  journal={arXiv preprint arXiv:2505.09388},
  year={2025}
}

@article{guo2025deepseek,
  title={Deepseek-r1: Incentivizing reasoning capability in llms via reinforcement learning},
  author={Guo, Daya and Yang, Dejian and Zhang, Haowei and Song, Junxiao and Zhang, Ruoyu and Xu, Runxin and Zhu, Qihao and Ma, Shirong and Wang, Peiyi and Bi, Xiao and others},
  journal={arXiv preprint arXiv:2501.12948},
  year={2025}
}

@article{chen2024huatuogpt-o1,
  title={Huatuogpt-o1, towards medical complex reasoning with llms},
  author={Chen, Junying and Cai, Zhenyang and Ji, Ke and Wang, Xidong and Liu, Wanlong and Wang, Rongsheng and Hou, Jianye and Wang, Benyou},
  journal={arXiv preprint arXiv:2412.18925},
  year={2024}
}

@article{shi2025openbioLLM,
  title={Fine-Tuning a Personalized OpenBioLLM Using Offline Reinforcement Learning.},
  author={Shi, Jinsheng and Yuan, Yuyu and Wang, Ao and Nie, Meng},
  journal={Applied Sciences (2076-3417)},
  volume={15},
  number={5},
  year={2025}
}

@article{wang2025beyond,
  title={Beyond the 80/20 rule: High-entropy minority tokens drive effective reinforcement learning for llm reasoning},
  author={Wang, Shenzhi and Yu, Le and Gao, Chang and Zheng, Chujie and Liu, Shixuan and Lu, Rui and Dang, Kai and Chen, Xionghui and Yang, Jianxin and Zhang, Zhenru and others},
  journal={arXiv preprint arXiv:2506.01939},
  year={2025}
}

@article{xu2025kdrl,
  title={KDRL: Post-Training Reasoning LLMs via Unified Knowledge Distillation and Reinforcement Learning},
  author={Xu, Hongling and Zhu, Qi and Deng, Heyuan and Li, Jinpeng and Hou, Lu and Wang, Yasheng and Shang, Lifeng and Xu, Ruifeng and Mi, Fei},
  journal={arXiv preprint arXiv:2506.02208},
  year={2025}
}

@inproceedings{zheng-etal-2024-llamafactory,
    title = "{L}lama{F}actory: Unified Efficient Fine-Tuning of 100+ Language Models",
    author = "Zheng, Yaowei  and
      Zhang, Richong  and
      Zhang, Junhao  and
      Ye, Yanhan  and
      Luo, Zheyan",
    editor = "Cao, Yixin  and
      Feng, Yang  and
      Xiong, Deyi",
    booktitle = "Proceedings of the 62nd Annual Meeting of the Association for Computational Linguistics (Volume 3: System Demonstrations)",
    month = aug,
    year = "2024",
    address = "Bangkok, Thailand",
    publisher = "Association for Computational Linguistics",
    url = "https://aclanthology.org/2024.acl-demos.38/",
    doi = "10.18653/v1/2024.acl-demos.38",
    pages = "400--410"
}

@inproceedings{hong2024orpo,
  title={ORPO: Monolithic Preference Optimization without Reference Model},
  author={Hong, Jiwoo and Lee, Noah and Thorne, James},
  booktitle={Proceedings of the 2024 Conference on Empirical Methods in Natural Language Processing},
  pages={11170--11189},
  year={2024}
}

@article{liao2025learnat,
  title={LearNAT: Learning NL2SQL with AST-guided Task Decomposition for Large Language Models},
  author={Liao, Weibin and Gao, Xin and Jia, Tianyu and Qiu, Rihong and Zhu, Yifan and Lin, Yang and Chu, Xu and Zhao, Junfeng and Wang, Yasha},
  journal={arXiv preprint arXiv:2504.02327},
  year={2025}
}

@inproceedings{liao2025tpo,
  title={TPO: Aligning Large Language Models with Multi-branch \& Multi-step Preference Trees},
  author={Liao, Weibin and Chu, Xu and Wang, Yasha},
  booktitle={The Thirteenth International Conference on Learning Representations},
  year={2025}
}

@inproceedings{liaomagical,
  title={Magical: Medical Lay Language Generation via Semantic Invariance and Layperson-tailored Adaptation},
  author={Liao, Weibin and Wang, Tianlong and Zhu, Yinghao and Wang, Yasha and Gao, Junyi and Ma, Liantao},
  booktitle={The Thirty-ninth Annual Conference on Neural Information Processing Systems}
}

@book{kurtz2017teaching,
  title={Teaching and learning communication skills in medicine},
  author={Kurtz, Suzanne and Draper, Juliet and Silverman, Jonathan},
  year={2017},
  publisher={CRC press}
}

@article{kurtz1996calgary,
  title={The Calgary—Cambridge Referenced Observation Guides: an aid to defining the curriculum and organizing the teaching in communication training programmes},
  author={Kurtz, Suzanne M and Silverman, Jonathan D},
  journal={Medical education},
  volume={30},
  number={2},
  pages={83--89},
  year={1996},
  publisher={Wiley Online Library}
}

@book{bickley2012bates,
  title={Bates' guide to physical examination and history-taking},
  author={Bickley, Lynn and Szilagyi, Peter G},
  year={2012},
  publisher={Lippincott Williams \& Wilkins}
}

\clearpage
\newpage
\appendix

\newpage

\section{Related Work}
\label{appendix:related_work}
\noindent\textbf{LLMs for Interactive Medical Questioning.}
Efforts have aimed to equip LLMs for multi-turn medical consultations. MMD-eval~\cite{liu2025MMD-eval} simulates doctor-patient interactions to evaluate LLMs. MEDIQ~\cite{li2024mediq} designs an abstention module and UoT~\cite{hu2024UoT} selects best questions according to simulated entropy-reduction. Others~\cite{liu2025dialogueT, liao2023automatic_consultation} fine-tune LLMs on constructed dialogues. However, these approaches rely on backbone capacity or static data, failing to truly enhance LLMs' proactive ability in dynamic dialogues. Question evaluation methods often rely on LLMs scored heuristics like usefulness or relevance~\cite{wang2025healthq,liaomagical} or leave-one-out estimations~\cite{hu2024UoT,lee2025good-utility-estimation,zhu2025ask-patient,mazzaccara2024EIG}, which fail to rigorously quantify questions’ values in complex medical contexts. Overall, there remains a lack of training frameworks that accurately assess question and optimize for LLMs' dynamic proactive ability.

\noindent\textbf{Reinforcement Learning for LLMs.}
Reinforcement learning (RL)~\cite{sutton1999reinforcement,kaelbling1996reinforcement} has proven effective for enhancing LLMs. RLHF~\cite{OpenAI2023GPT4TR,kaufmann2024survey-RLHF} aligns models via reward modeling and PPO~\cite{schulman2017ppo}. DPO~\cite{rafailov2023dpo} bypasses explicit reward modeling by learning from preference pairs. GRPO~\cite{shao2024deepseekmath-grpo} and DAPO~\cite{yu2025dapo} leverage group-wise advantages for optimization and enhance reasoning abilities. However, RL for interactive medical consultations with tailored rewards remains underexplored.

\section{Patient Simulator}
\label{appendix:patient_simulator}

To enable scalable training and evaluation for interactive medical consultation, we implement an LLM-based patient simulator. While such simulation reduces interaction costs, it may introduce biases stemming from the backbone model, response control mechanisms, or behavioral styles. To ensure that \M~does not depend on a specific simulator instantiation, we systematically design and evaluate multiple patient simulator variants.

\begin{figure*}[t]
    \centering
    \includegraphics[width=\linewidth]{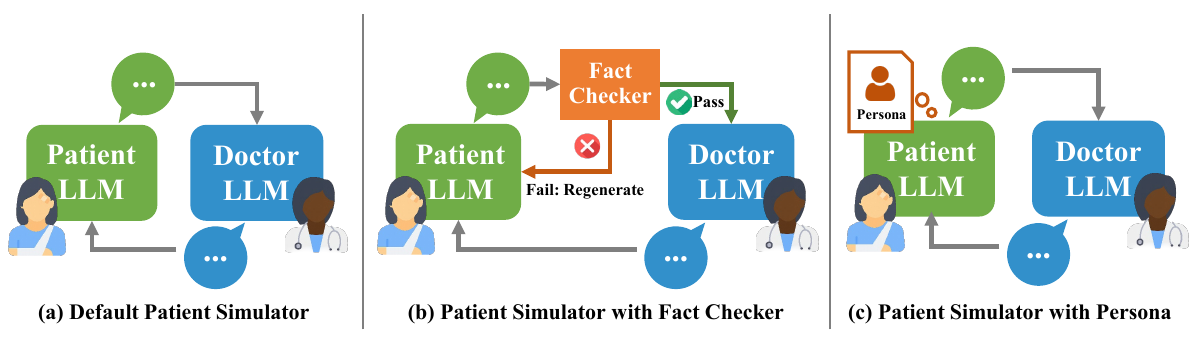}
    \caption{Illustration of patient simulator designs. 
    \textbf{(a) Default}: the simulator directly responds to questions. 
    \textbf{(b) Fact checker}: a fact checker module triggers regeneration of responses to ensure factuality and relevance.
    \textbf{(c) Persona}: the simulator is conditioned on randomized MBTI persona traits to induce diverse communication behaviors.}
    \label{fig:patient_simulator_design}
\end{figure*}

\noindent\textbf{Simulator Designs.} Specifically, we consider simulator designs along three dimensions:
(1) the backbone LLM used to generate patient responses, which determines the linguistic capacity and medical knowledge of the simulator;
(2) the interaction process, including whether question quality is explicitly validated; and
(3) the behavioral variability introduced via persona modeling.
Figure~\ref{fig:patient_simulator_design} provides an overview of the three representative simulator designs evaluated in this work.

\begin{itemize}[leftmargin=*,noitemsep,topsep=2pt]
    \item \textbf{Default Patient Simulator.}
In the default setting, the patient simulator directly answers the doctor LLM’s questions based on the complete fact set $\mathcal{F}$. Given a question, the simulator generates a response grounded in $\mathcal{F}$ whenever possible; otherwise, it replies ``I don’t know.''  
Under this setting, we experiment with multiple backbone models, including \textit{Qwen2.5-72B-Instruct}, \textit{Qwen3-32B}, \textit{GPT-5}, and \textit{DeepSeek-V3.2}, to examine the impact of backbone choice.

\item \textbf{Patient Simulator with Fact Checker.}
To improve response reliability, we introduce a fact-checking module that explicitly evaluates each generated patient response along two dimensions: relevance, i.e., whether the response addresses the doctor’s question, and factuality, i.e., whether the response is supported by the atomic fact set $\mathcal{F}$.
If either criterion is violated, the response is rejected and the simulator is prompted to regenerate, with explicit feedback indicating the detected issue (e.g., irrelevance or factual inconsistency). This process is repeated until both criteria are satisfied. Such a design enforces strict quality control of the patient simulator.

\item \textbf{Patient Simulator with Persona.}
To model realistic variability in patient communication, we construct a persona-based simulator by conditioning the backbone LLM on randomized MBTI persona profiles. Each profile provides a structured description of the patient’s personality traits, together with explicit guidance on corresponding communication styles, such as verbosity, tone, and level of certainty. The simulator is instructed to consistently adhere to the assigned persona when responding to the doctor’s questions, resulting in diverse yet coherent interaction patterns (e.g., concise and cautious responses versus detailed and expressive ones), while still grounding answers in the atomic fact set $\mathcal{F}$.
\end{itemize}

\begin{table}[t]
\centering
\small
\resizebox{\linewidth}{!}
{
\begin{tabular}{lcc}
\toprule
\rowcolor[gray]{0.95}
\textbf{Patient Simulator Setting} & \textbf{Relevance (\%)} & \textbf{Factuality (\%)} \\
\midrule
\multicolumn{3}{l}{\emph{Backbone Replacement (Default Simulator)}} \\
\quad \textsc{Qwen2.5-72B}           & \textbf{96.80} & 92.40 \\
\quad \textsc{Qwen3-32B}             & 95.60 & 92.40 \\
\quad \textsc{DeepSeek-V3.2}        & 93.60 & 95.60 \\
\quad \textsc{GPT-5}                & \underline{96.40} & \underline{96.80} \\
\midrule
\multicolumn{3}{l}{\emph{Enhanced Simulation Mechanisms}} \\
\quad \textsc{Qwen3-32B} + Fact-checker & 94.40 & \textbf{97.20} \\
\quad \textsc{Qwen3-32B} + MBTI persona & 95.20 & 94.40 \\
\midrule
\multicolumn{3}{l}{\emph{Human Verification Result}} \\
\quad \textsc{Qwen2.5-72B} & 96.00 & 92.00 \\
\bottomrule
\end{tabular}
}
\caption{Quality evaluation of patient simulator configurations. Relevance measures whether the simulated patient responses address the doctor’s questions, while factuality measures consistency with the provided facts.}
\label{tab:patient_simulator_quality}
\end{table}

\noindent\textbf{Automatic Evaluation.}
We evaluate simulator response quality along two dimensions: \textbf{Relevance}, measuring whether the response meaningfully addresses the doctor’s question, and \textbf{Factuality}, measuring whether the response is supported by the atomic facts. Both criteria are assessed using an LLM-based evaluator (based on DeepSeek-V3.2~\cite{deepseekai2025deepseekv32}) with binary (Yes/No) judgments.
For each patient simulator configuration, we randomly sample 250 interactions from both the CMB and MedQA benchmarks, resulting in 500 evaluated interactions per simulator. Table~\ref{tab:patient_simulator_quality} summarizes the relevance and factuality accuracy across different simulator configurations. 
Overall, LLM-based patient simulators demonstrate consistently high relevance and factuality, making them reliable participants in interactive medical consultations for model training and evaluation. Fact checker improves factuality at a small cost to relevance, while persona modeling introduces mild behavioral variability without degrading overall response quality. For simplicity, we implement default patient simulator using Qwen2.5-72B, which achieves a strong balance between response quality and computational cost.

\noindent\textbf{Human Verification.}
To further validate the reliability of our chosen simulator, we manually annotate a small subset (50 samples) of interactions generated by the Qwen2.5-72B-based patient. The human evaluation closely matches the automatic assessment, confirming that the simulator provides coherent, relevant, and factually grounded responses suitable for interactive medical consultation experiments.

\section{Supplementary Experiments}
\label{appendix:supplementary_experiments}

\begin{table}[t]
\centering
\resizebox{\columnwidth}{!}{
\begin{tabular}{lcc}
\toprule
\rowcolor[gray]{0.95}
\textbf{Model} 
& \makecell{\textbf{Outcome} \\ \textbf{Prediction (\%)}} 
& \makecell{\textbf{Readmission} \\ \textbf{Prediction (\%)}} \\
\midrule
Vanilla & 56.90 & 41.60 \\
ProMed(S\#1)+GRPO & 59.90 & 50.40 \\
ProMed(S\#1+2) & \textbf{69.50} & \textbf{76.20} \\
\bottomrule
\end{tabular}
}
\caption{Performance on MIMIC-IV prediction tasks.}
\label{tab:mimic_results}
\end{table}

\subsection{Experiments on MIMIC-IV}
\label{appendix:mimic4}
While our primary evaluation utilizes natural-language dialogue to simulate the most common mode of doctor--patient interaction, real-world clinical intelligence must handle heterogeneous data sources. To further validate this capability, we conduct additional experiments on the MIMIC-IV~\cite{johnson2023mimic} dataset, demonstrating that our model can proactively acquire and distinguish critical information from uncurated, noisy, and mixed structured--unstructured medical data.

\paragraph{Data.}
We randomly sample 1000 patient records from MIMIC-IV. Each record contains demographic information (e.g., age and gender) and multiple hospital admissions. Each admission includes both structured laboratory measurements (numerical values) and unstructured clinical notes.

\paragraph{Information Sufficiency Setting.}
To simulate partial observability: The doctor LLM initially observes patient demographics, 50\% randomly sampled laboratory tests, and the first 500 characters of clinical notes. Patient Simulator has access to the full patient record. Neither side has access to the ground-truth prediction target, preventing information leakage and the model trivially asking for the answer.

\paragraph{Tasks.}
We evaluate two clinically relevant prediction tasks:
\begin{itemize}[leftmargin=*,noitemsep,topsep=2pt]
    \item \textbf{Mortality outcome prediction}: Predict patient outcome (mortality) from longitudinal visit sequences.
    \item \textbf{Readmission prediction}: Predict whether a patient will have a subsequent admission based on a single visit.
\end{itemize}
Structured data are serialized into textual form with explicit column names (e.g., \textit{Heart Rate: 90.0}) and concatenated with clinical notes as model input.

\paragraph{Models.}
LLaMA-3.1-8B-Instruct is the vanilla backbone model. \M(S\#1+GRPO) is the GRPO baseline trained with answer-correctness rewards. \M(S\#1+2) is trained with our framework. Training is conducted on the CMB dataset.

\paragraph{Results.} As shown in table~\ref{tab:mimic_results}, \M~significantly outperforms all baselines on both tasks. The untrained backbone model performs near or below random guessing, highlighting the difficulty of reasoning over noisy and heterogeneous clinical data. 
The GRPO-trained model shows modest improvement, reflecting slightly enhanced medical abilities; however, because its RL objective optimizes only final answer correctness, it fails to consistently acquire critical intermediate information in this OOD setting. In contrast, \M~explicitly rewards high-value questions, which enables the model to actively acquire and extract task-relevant signals from mixed structured and unstructured data.

\paragraph{Case Study Analysis.} Further qualitative analysis (see Appendix~\ref{appendix:case_study} for details) reveals:
\begin{itemize}[leftmargin=*,noitemsep,topsep=2pt]
    \item In the \textbf{outcome prediction} case: The GRPO model is misled by superficially positive signals (e.g., stable vitals). In contrast, \M~retrieves high-impact factors such as respiratory failure, multimorbidity, and functional decline, leading to the correct prediction.
    \item In the \textbf{readmission prediction} case: The GRPO model focuses narrowly on ongoing medical issues, resulting in an incorrect prediction. In contrast, \M~performs more comprehensive information retrieval across clinical and care-related dimensions (e.g., functional status, caregiver support, hospice care), enabling the correct decision.
\end{itemize}

\paragraph{Summary.}
These findings suggest that our training strategy equips the model with the ability to handle noisy, partially observed, and heterogeneous medical data, which is essential for real-world clinical deployment.

\subsection{Clinical SOP-Guided Questioning}
\label{sec:appendix_sop}

\begin{table}[t]
\centering
\small
\resizebox{\columnwidth}{!}{
\begin{tabular}{llccc}
\toprule
\rowcolor[gray]{0.95}
\textbf{Dataset} & \textbf{Model} & \textbf{Prompt} & \textbf{Accuracy (\%)} & \textbf{Avg. \#Questions} \\
\midrule
CMB   & LLaMA3-8B & Original    & \textbf{44.49} & 4.13 \\
CMB   & LLaMA3-8B & SOP-guided  & 43.30          & 4.48 \\
CMB   & ProMed    & Original    & \textbf{59.33} & 4.40 \\
CMB   & ProMed    & SOP-guided  & 52.75          & 3.67 \\
\midrule
MedQA & LLaMA3-8B & Original    & 40.11          & 4.04 \\
MedQA & LLaMA3-8B & SOP-guided  & \textbf{47.10} & 5.66 \\
MedQA & ProMed    & Original    & \textbf{57.50} & 3.70 \\
MedQA & ProMed    & SOP-guided  & 52.42          & 5.74 \\
\bottomrule
\end{tabular}
}
\caption{Impact of SOP-guided questioning on accuracy and efficiency.}
\label{tab:sop_results}
\end{table}

Question ordering plays an important role in clinical reasoning, potentially affecting both diagnostic efficiency and accuracy. To systematically study this factor, we conduct experiments incorporating an explicit clinical questioning protocol into the model prompting.

\paragraph{Clinical Inquiry Protocol.}
We introduce a \textbf{Clinical Inquiry Protocol} into the system prompt, requiring the doctor model to follow the Calgary--Cambridge Framework~\cite{kurtz1996calgary,kurtz2017teaching}, a widely adopted standard for medical interviewing. The protocol enforces a structured questioning order: (1) symptom clarification, following the OPQRST scheme~\cite{bickley2012bates}, (2) past medical history, (3) social and family history, and (4) review of systems.

\paragraph{Experimental Setup.}
We compare our original prompting and SOP-guided prompting across the backbone model and our trained \M~model, evaluated on CMB (in-domain) and MedQA (OOD). We report final accuracy and average number of questions asked (as a proxy for efficiency).

\paragraph{Results.}
\textbf{Model- and context-dependent effectiveness.}
The impact of SOP guidance varies across models and datasets. On the CMB dataset, enforcing the protocol degrades performance for both models. On MedQA, improvements are observed only for the untrained backbone model, while ProMed experiences a performance drop.

\textbf{Efficiency trade-offs.}
SOP guidance generally increases the number of questions asked (in three out of four settings), indicating reduced efficiency. Even when accuracy improves (e.g., LLaMA3-8B on MedQA), this gain comes at the cost of additional queries.

\textbf{Interference with learned policies.}
For the ProMed model, which has already learned effective information-seeking strategies, imposing a rigid protocol consistently harms performance. This suggests a mismatch between fixed procedural constraints and the model's learned questioning policy.

\textbf{Summary.}
These findings indicate that while structured questioning is clinically meaningful, a single fixed SOP does not consistently improve performance across tasks or domains. In practice, effective questioning strategies are often domain-specific and adaptive. Rigid protocols may therefore limit the model’s ability to optimize information acquisition. In contrast, approaches that prioritize information gain provide a more flexible and robust foundation for clinical reasoning.

\subsection{Comparison with Medical LLMs}

\begin{figure}[t]
    \centering
    \includegraphics[width=1\linewidth]{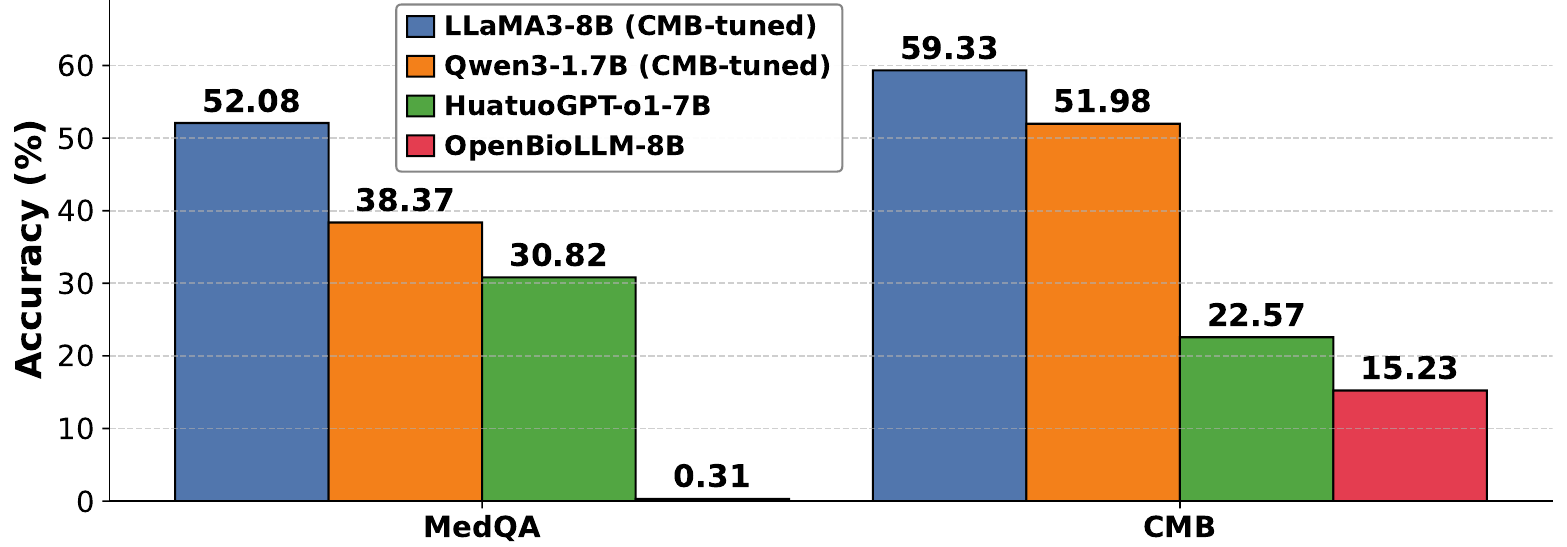}
    \caption{Performance comparison of \M{} with various SOTA medical LLMs.}
    \label{fig:comparison_with_medical_LLMs}
\end{figure}

To further validate the necessity of optimizing proactive information-seeking, we compare \M-optimized models (LLaMA3-8B and Qwen3-1.7B, trained on CMB) with open-source medical LLMs \textit{HuatuoGPT-o1-7B}~\cite{chen2024huatuogpt-o1} and \textit{OpenBioLLM-8B}~\cite{shi2025openbioLLM}, as shown in Figure~\ref{fig:comparison_with_medical_LLMs}. \textbf{\M~enables strong interactive reasoning beyond medical training.} Despite being of comparable or smaller scales, \M~models outperform existing medical LLMs on both benchmarks by large margins, demonstrating that medical pre-training and SFT alone does not endow LLMs with robust interactive diagnostic abilities. These findings confirm that targeted training is essential for enabling clinically valuable interaction.

\subsection{Comparison with proprietary LLMs}

\begin{figure}[t]
    \centering
    \includegraphics[width=\linewidth]{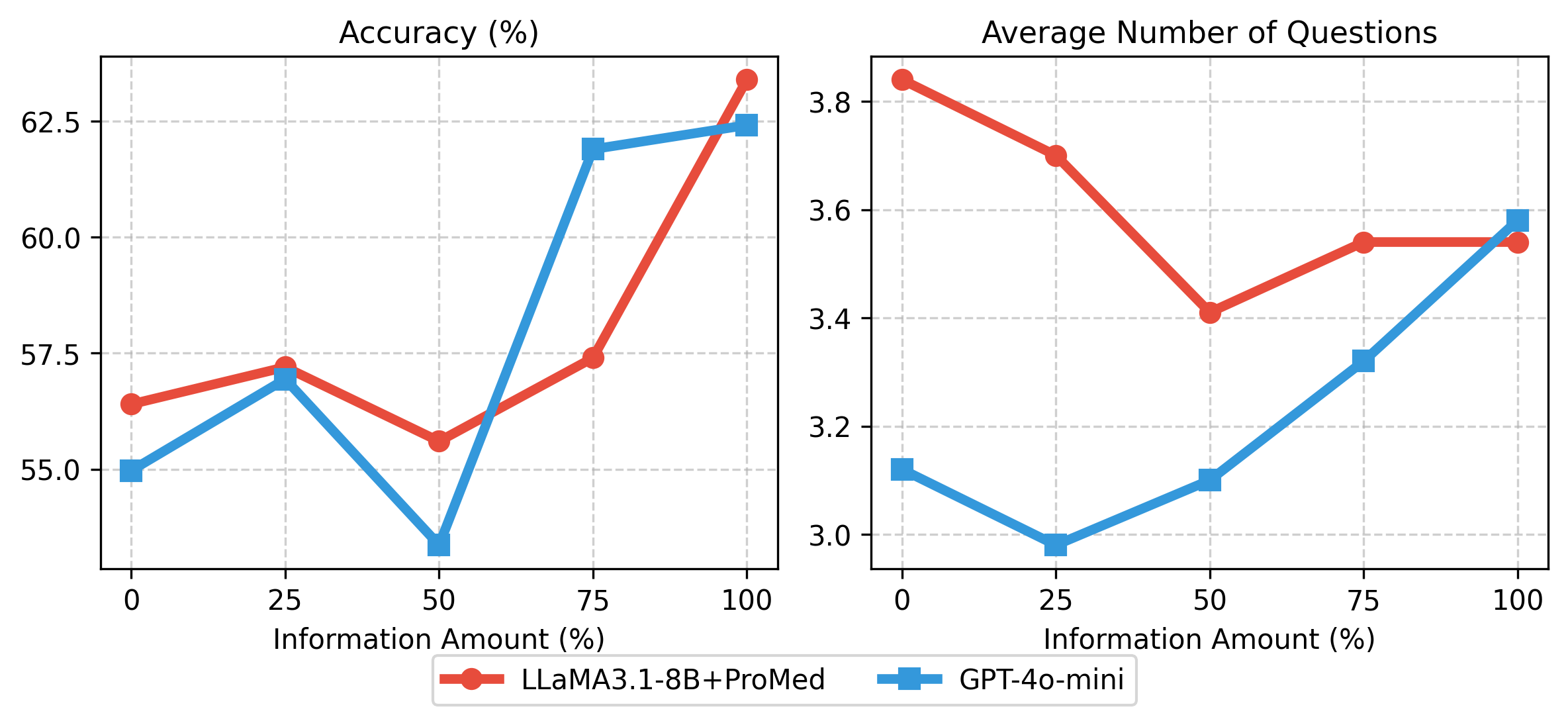}
    \caption{Comparison with GPT-4o-mini.}
    \label{fig:gpt_comparison}
\end{figure}

To further evaluate the effectiveness of \M, we compare our fine-tuned LLaMA3.1-8B against the proprietary GPT-4o-mini, across varying levels of initial information availability. As shown in Figure \ref{fig:gpt_comparison}, our model consistently matches or outperforms GPT-4o-mini in accuracy. This suggests that \M~training successfully empowers an open-sourced model to reach, and even exceed, the reasoning capabilities of proprietary counterparts in specialized medical domains. The distinction also lies in the models' adaptive questioning behavior: 
\begin{itemize}[leftmargin=*,noitemsep,topsep=2pt]
\item \textbf{ProMed Efficiency:} Our model exhibits a highly adaptive inquiry strategy. It intelligently reduces the number of questions as the initial information becomes more sufficient (from 0\% to 50\%). 
\item \textbf{GPT-4o-mini Redundancy:} In contrast, GPT-4o-mini shows an inefficient trend; its average number of questions actually \textit{increases} with more available information. This indicates a lack of awareness regarding information sufficiency, leading to redundant queries even when the evidence is already conclusive.
\end{itemize}
These results underscore that \M~instills a "self-awareness" of information gaps. By precisely quantifying the utility of questions through SIG rewards, our model learns to prioritize high-gain inquiries and refrain from redundant dialogue, achieving a more professional and human-like clinical reasoning process than general-purpose proprietary models.

\begin{figure}[t]
    \centering
    \includegraphics[width=0.7\linewidth]{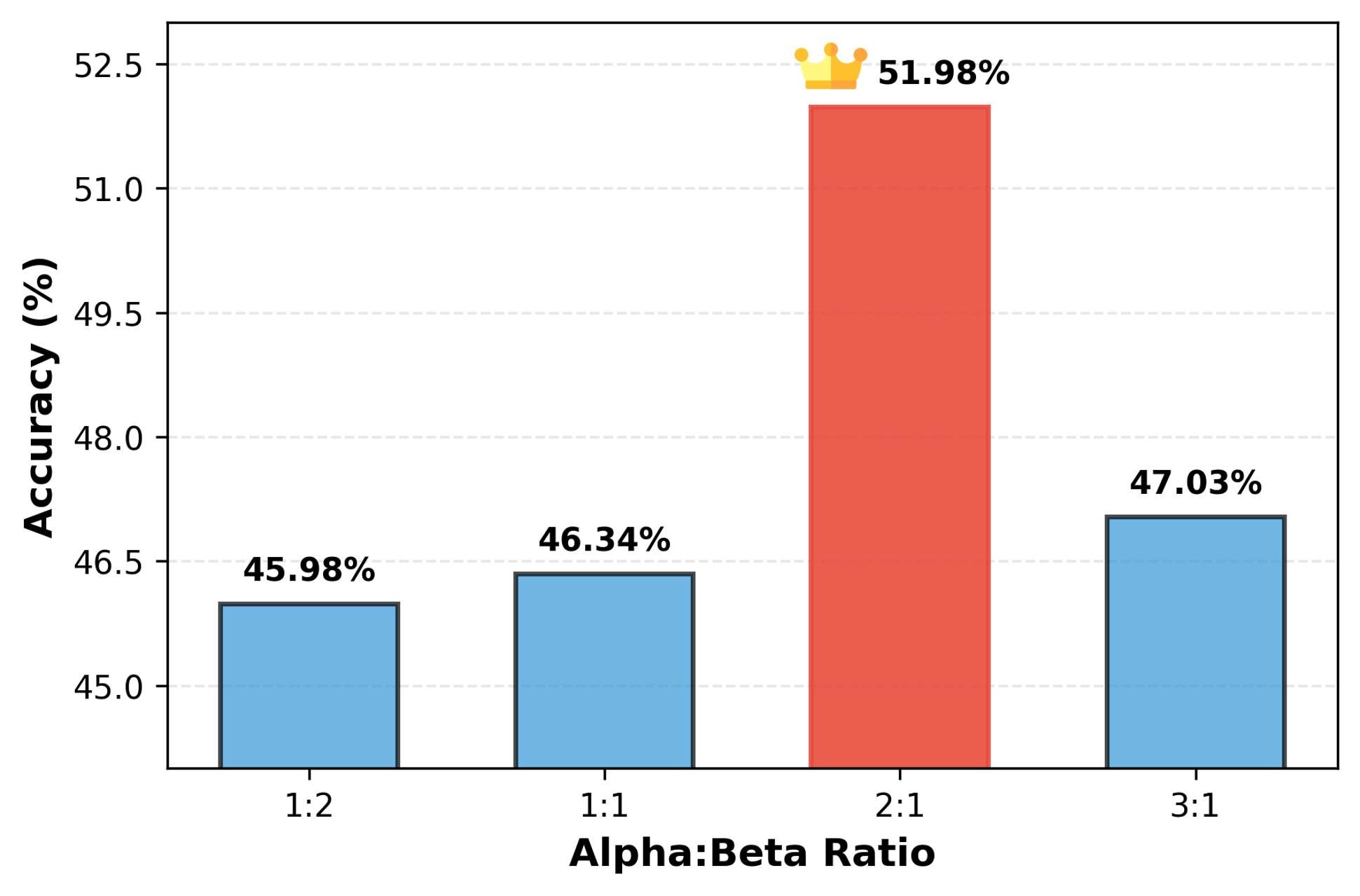}
    \caption{Sensitivity analysis of the reward weights $\alpha$ and $\beta$ on the CMB dataset using Qwen3-1.7B.}
    \label{fig:alpha_beta_sensitivity}
\end{figure}

\subsection{Hyperparameter Sensitivity Analysis}

To investigate the impact of the reward components on model performance, we conduct a sensitivity analysis on the weighting coefficients $\alpha$ (answer outcome reward) and $\beta$ (question process reward). We evaluate four different ratios: $\alpha:\beta \in \{1:1, 2:1, 3:1, 1:2\}$ by training Qwen3-1.7B on CMB. As illustrated in Figure~\ref{fig:alpha_beta_sensitivity}, Our primary configuration ($\alpha:\beta = 2:1$) achieves the highest accuracy of 51.98\%. When the outcome reward is further emphasized, performance decreases, indicating that overly prioritizing answer correctness can weaken the model’s incentive to ask informative questions during the interaction process. On the other hand, increasing question process reward weight also degrades performance, implying that excessive focus on question quality without sufficient guidance from outcome supervision may lead to suboptimal diagnostic decisions. Overall, these results highlight the importance of a balanced reward design. A moderate emphasis on the outcome reward, while still preserving a strong signal for question quality, yields the best performance. This observation supports our design choice of $\alpha:\beta=2:1$ and demonstrates that effective interactive medical reasoning requires both accurate final predictions and high-quality information-seeking behavior.

\subsection{Information Shapley Analysis}
To further validate the effectiveness of Shapley values in capturing clinically meaningful information, we design a noise-injection experiment. Specifically, we randomly sample 100 patient cases in MedQA and inject irrelevant facts, such as “the patient’s zodiac sign” or “the patient’s hair color” that are unrelated to clinical outcomes. We then compute the importance scores of all atomic facts using two methods: Leave-One-Out (LOO) and our proposed atomic fact Shapley. In LOO, the importance of each fact is measured individually by adding it to the input and observing its impact on the model's probability to predict the correct answer. For each method, we rank all facts by their estimated importance and calculate Recall@K, which measures the proportion of truly relevant medical facts appearing in the top-K positions. 

As shown in Table~\ref{tab:shapley_vs_loo}, Shapley values consistently outperform LOO across all K. Notably, Shapley achieves a Recall@1 of 95.96\% and Recall@3 of 90.24\%, compared to LOO’s 80.16\% and 76.06\%, respectively. These results demonstrate that Shapley values provide a finer-grained and more accurate reflection of clinical relevance than LOO, enabling more reliable identification of medically salient information. This superior sensitivity to fact-level importance supports the use of Shapley values as the foundation of our reward design in interactive medical questioning.

\begin{table}[t]
\setlength{\tabcolsep}{3pt}
\centering
\small
\begin{tabular}{lcccc}
\toprule
\rowcolor[gray]{0.95}
\textbf{Method} & \textbf{Recall@1} & \textbf{Recall@3}  & \textbf{Recall@10} \\
\midrule
\textbf{Shapley} & \textbf{0.9596} & \textbf{0.9024} & \textbf{0.9058} \\
\textbf{LOO} & 0.8016 & 0.7606 & 0.8275 \\
\bottomrule
\end{tabular}
\caption{Comparison of Recall@K between Shapley Value and Leave-One-Out (LOO) under noise injection setting. Higher values indicate better ability to identify clinically relevant facts.}
\label{tab:shapley_vs_loo}
\end{table}

\section{Monte Carlo Shapley}
\label{appendix:monte_carlo_shapley}
\subsection{Algorithm Details}
To mitigate the computational overhead of Monte Carlo Shapley calculation, we implement a batched inference strategy leveraging the vLLM~\cite{kwon2023efficient} framework. For each sampled permutation $\pi_k$, we construct a sequence of $n$ prefixes, $\{S_1, \dots, S_n\}$, representing the cumulative addition of facts. Rather than performing $n$ independent sequential passes, we pack these prefixes into a single inference batch. Specifically, we utilize the prompt logprobs feature of the inference engine to extract the log-probabilities of the target answer $A$ conditioned on each prefix. This batching approach reduces the number of model forward passes from $O(K \times n)$ to $O(K)$, where $K$ is the number of MC iterations and $n$ is the number of atomic facts.

In Algorithm~\ref{alg:mcshapley}, we provide the pseudo-codes of the batched Monte Carlo approximation algorithm for calculating the atomic fact Shapley in Section~\ref{sec:SIG_Reward}.

\begin{algorithm}[t]
\caption{Batched Monte Carlo Fact Shapley}
\label{alg:mcshapley}
\begin{algorithmic}[1]
\Require Model $M$, Atomic Facts $\mathcal{F}=\{f_1,\dots,f_n\}$, Question $Q$, Answer $A$, Max iterations $K$, Tolerance $\epsilon$
\Ensure Estimated Shapley values $\{\phi(f_i)\}_{i=1}^n$

\State Initialize $\phi(f_i) \gets 0$ for all $i$
\State Compute $v(S_0) \gets \log P_M(A\mid \emptyset,Q)$

\For{$k=1$ to $K$}
    \State Sample permutation $\pi_k$ of $\{1,\dots,n\}$
    \State Construct fact subsets $S_j=\{f_{\pi(1)},\dots,f_{\pi(j)}\}$
    \State \textbf{Batch compute} $\{v(S_1),\dots,v(S_n)\}$ via parallel inference
    \State $v_{\text{prev}} \gets v(S_0)$

    \For{$j=1$ to $n$}
        \State $i \gets \pi(j)$
        \State $\Delta \gets v(S_j) - v_{\text{prev}}$
        \State $\phi(f_i)\gets \frac{k-1}{k}\phi(f_i)+\frac{1}{k}\Delta$
        \State $v_{\text{prev}}\gets v(S_j)$
        \If{$|v(S_j)-v_{\mathcal{F}}|<\epsilon$}
            \State \textbf{break}
        \EndIf
    \EndFor

    \If{$\frac{1}{k-1}\sum_i |\phi_i^{(k)}-\phi_i^{(k-1)}|<\epsilon$}
        \State \textbf{break}  \Comment{Convergence of estimation}
    \EndIf
\EndFor
\State \Return $\{\phi(f_i)\}_{i=1}^n$
\end{algorithmic}
\end{algorithm}

\subsection{Complexity Analysis}
\begin{figure*}[t]
    \centering
    \includegraphics[width=\linewidth]{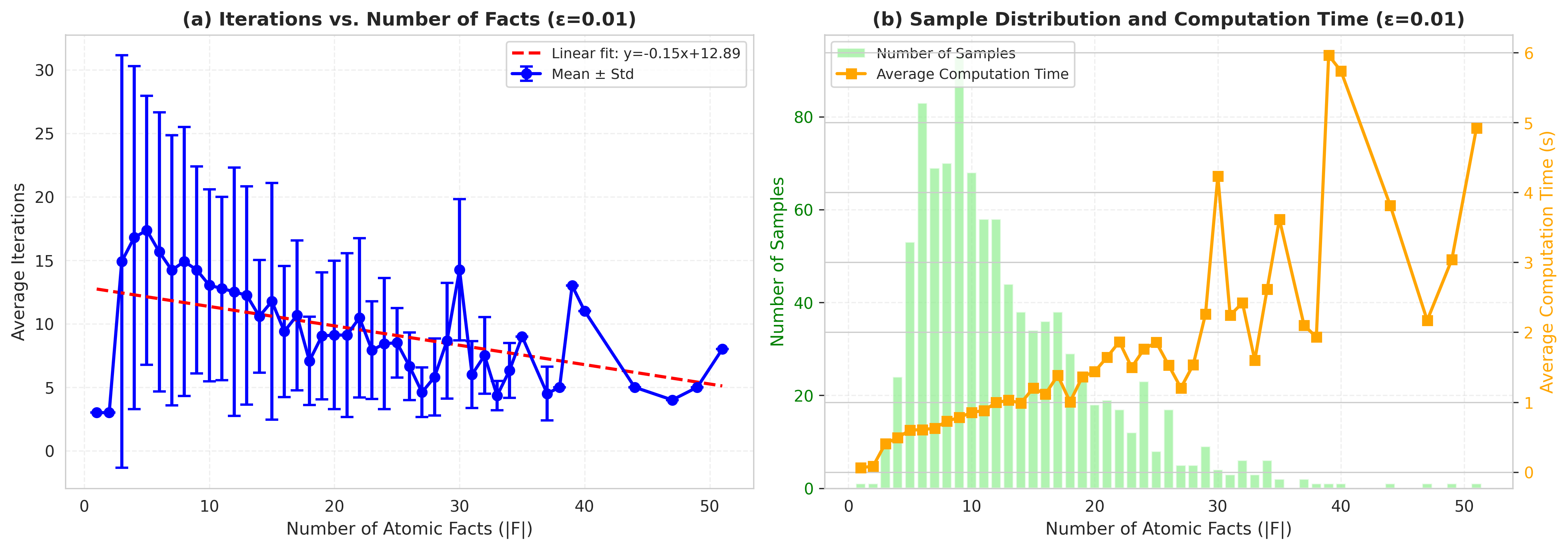}
    \caption{Efficiency analysis of Atomic Fact Shapley calculation. (a) Average iterations required for convergence vs. number of facts; (b) Distribution of samples and average computation time per fact set size.}
    \label{fig:efficiency_analysis}
\end{figure*}

\begin{table}[t]
\centering
\small
\begin{tabular}{ccc}
\toprule
\rowcolor[gray]{0.95}
\textbf{Tolerance ($\epsilon$)} & \textbf{Average Time (s)} & \textbf{Std Dev} \\ 
\midrule
0.10 & 0.34 & 0.25 \\
0.01 & 1.03 & 0.82 \\
0.001 & 3.69 & 2.91 \\
\bottomrule
\end{tabular}
\caption{Impact of Convergence Tolerance ($\epsilon$) on Computation Time.}
\label{tab:tolerance}
\end{table}

\begin{table}[t]
\centering
\small
\begin{tabular}{ccc}
\toprule
\rowcolor[gray]{0.95}
\textbf{Model Size (B)} & \textbf{Average Time (s)} & \textbf{Std Dev} \\ 
\midrule
1.7 & 1.03 & 0.82 \\
3.0 & 6.27 & 5.18 \\
8.0 & 10.47 & 8.84 \\ 
\bottomrule
\end{tabular}
\caption{Impact of Model Scale on Time Complexity.}
\label{tab:modelsize}
\end{table}

We analyze the SIG reward's efficiency, focusing on how the computational cost scales with information volume (number of atomic facts), precision requirements (convergence tolerance), and model capacity. To ensure a representative evaluation, we randomly sampled 500 instances each from the training sets of CMB and MedQA for the tests. 

\noindent\textbf{Impact of the Number of Atomic Facts ($\mathcal{F}$).} As illustrated in Figure~\ref{fig:efficiency_analysis}, the computational cost of the SIG reward remains efficient even as the number of atomic facts increases. While the theoretical search space for Shapley values is $2^{|\mathcal{F}|}$, our batch-processing MC estimation reduces the empirical complexity. Specifically, Fig. (a) shows that the number of iterations required for convergence stays relatively stable (between 5 to 15 iterations) and even exhibits a slight downward trend as $|\mathcal{F}|$ increases. This suggests that in larger medical evidence sets, the marginal contributions often converge quickly. Fig. (b) confirms that the average computation time scales linearly with $|\mathcal{F}|$, typically remaining under 2 seconds. The spikes in computation time at higher $|\mathcal{F}|$ values are largely due to the increased prompt length and the linear growth of prefix evaluations within each batch. Moreover, as shown by the sample distribution in Figure (b), the number of facts per query rarely exceeds 30 in both CMB and MedQA datasets. This implies that high-latency cases are infrequent in practice, with most estimations concluding in under 2 seconds.

\noindent\textbf{Impact of Convergence Tolerance ($\epsilon$).} The choice of tolerance $\epsilon$ presents a direct trade-off between estimation precision and latency. As shown in Table~\ref{tab:tolerance}, setting $\epsilon = 0.00$ (full iteration) results in an average latency of 3.69s. However, by relaxing $\epsilon$ to 0.01, we achieve a 3.6$\times$ speedup (1.03s) with negligible impact on reward quality. A further increase to $\epsilon=0.10$ reduces time to 0.34s, making it suitable for real-time interactive scenarios.

\noindent\textbf{Impact of Model Scale.} Table~\ref{tab:modelsize} details the impact of the model's parameter size on efficiency. While the 1.7B model is highly efficient (1.03s), moving to 8B parameters increases the latency. However, this growth remains near-linear relative to the model size. This scaling confirms that the time complexity is predictable and manageable, further underscoring the necessity of our batched inference strategy, which ensures that larger, more capable models can be integrated without incurring prohibitive, non-linear delays.

Overall, the results demonstrate that our batched Monte Carlo estimation of fact Shapley maintains high practical scalability. The detailed runtime analysis of integrating this reward into the overall framework is provided in Appendix~\ref{sec:framework_complexity}.

\section{SIG-Guided MCTS Sampling}
\label{appendix:mcts_sampling}
\begin{algorithm}[t]
\caption{SIG-Guided MCTS Sampling}
\label{alg:sig_mcts}
\begin{algorithmic}[1]
\Require Initial inquiry $Q_p$, atomic facts $\mathcal{F}$, ground-truth answer $A^*$,
model $\mathcal{M}$, simulated patient $\mathcal{P}_{\text{sim}}$,
maximum depth $T_{\max}$, number of simulations $N$
\Ensure Answer-correct optimal trajectory
$\tau^* = \{Q_p,(q_1,r_1),\dots,(q_T,r_T),A'\}$

\State Initialize root node $n_0 \gets Q_p$
\State Initialize best trajectory $\tau^* \gets \emptyset$, best reward $R^* \gets -\infty$

\For{$i=1$ to $N$}
    \State Initialize path $\mathcal{P} \gets [n_0]$, history $\mathcal{H}_0 \gets \emptyset$
    
    \Comment{Selection}
    \While{$n$ is fully expanded \textbf{and} not terminal}
        \State $n \gets \arg\max_{n'} \mathrm{UCT}(n')$
        \State $\mathcal{P} \gets \mathcal{P} \cup \{n\}$
    \EndWhile

    \Comment{Expansion}
    \If{$n$ is not terminal}
        \State Generate question $q_t \gets \mathcal{M}(Q_p,\mathcal{H}_{t-1})$
        \State Simulate response $r_t \gets \mathcal{P}_{\text{sim}}(q_t,\mathcal{F})$
        \State $\mathcal{H}_t \gets \mathcal{H}_{t-1} \cup \{(q_t,r_t)\}$
        \State Add node $n_t=(q_t,r_t)$ to $\mathcal{P}$
        \State Compute local reward $R_{\text{local}}(q_t) \gets \mathrm{SIG}(q_t)$
    \Else
        \State Predict final answer $A' \gets \mathcal{M}(Q_p,\mathcal{H}_{t-1})$
    \EndIf

    \Comment{Simulation}
    \While{not terminal \textbf{and} $t < T_{\max}$}
        \State Generate question $q_t$, simulate response $r_t$
        \State $\mathcal{H}_t \gets \mathcal{H}_{t-1} \cup \{(q_t,r_t)\}$
    \EndWhile
    \State Predict answer $A' \gets \mathcal{M}(Q_p,\mathcal{H}_t)$
    \State Compute trajectory reward $R(\tau)$

    \Comment{Backpropagation}
    \ForAll{$n \in \mathcal{P}$}
        \State $N(n) \gets N(n) + 1$
        \State $W(n) \gets W(n) + R(\tau)$
        \State $\bar{R}(n) \gets W(n)/N(n)$
    \EndFor

    \If{$A' = A^*$ \textbf{and} $R(\tau) > R^*$}
        \State $\tau^* \gets \tau$
        \State $R^* \gets R(\tau)$
    \EndIf
\EndFor

\State \Return $\tau^*$
\end{algorithmic}
\end{algorithm}

\subsection{Algorithm Details}
To construct high-quality SFT interaction trajectories, we employ the SIG-Guided MCTS~\cite{coulom2006MCTS} algorithm. Below, we provide detailed descriptions of the sampling process. Pseudo-codes for this process is provided in Algorithm~\ref{alg:sig_mcts}.

Each MCTS run simulates a doctor-patient dialogue tree rooted at the initial clinical inquiry \( Q_p \), where nodes represent interaction states. The search space is defined by the model’s question generation distribution \( \mathcal{M}(\cdot \mid \mathcal{H}_{t}) \), where \( \mathcal{H}_{t} = \{(q_1, r_1), ..., (q_t, r_t)\} \) is the dialogue history up to step \( t \).
The MCTS search follows the process:

\begin{itemize}[leftmargin=*,noitemsep,topsep=2pt]
    \item \textbf{Selection.} 
Starting from the root node $n_0$, the algorithm selects a child node $n'$ at each step that maximizes the Upper Confidence Bound for Trees (UCT)~\cite{kocsis2006UCT}:
\begin{equation}
\text{UCT}(n') = \bar{R}(n') + c \cdot \sqrt{\frac{\log N(n)}{N(n') + \epsilon}},
\end{equation}
where $\bar{R}(n')$ is the average total reward of node $n'$, $N(n)$ is the number of visits to parent node $n$, and $c$ is an exploration coefficient. This process continues recursively until a leaf or unexpanded node is reached.
    \item \textbf{Expansion.} 
    Given selected node \( n_{t-1} \), the model decides to either:
    \begin{enumerate}[label=(\alph*),nosep]
        \item Generate a follow-up question \( q_t \), receive a response \( r_t \) from a simulated patient, and form new node \( n_t = (q_t, r_t) \), or
        \item Issue a final answer \( A' \) and terminate.
    \end{enumerate}
    If expanded, we update:
    \begin{align}
        \mathcal{H}_t &= \mathcal{H}_{t-1} \cup \{(q_t, r_t)\}, \\
        U_t &= \mathcal{M}_\text{Understand}(Q_p,\mathcal{H}_t), \\
        R_{\text{local}}(q_t) &= \text{SIG}(q_t).
    \end{align}

    \item \textbf{Simulation.} 
    The interaction proceeds recursively until the model issues a final answer \( A' \) or reaches a depth limit \( T_{\text{max}} \). The trajectory reward is calculated as in Eq 7 in Section 3.3.

    \item \textbf{Backpropagation.} 
    The final reward \( R(\tau) \) is propagated to all nodes \( n \) along the selected path:
    \begin{align}
        N(n) &\gets N(n) + 1, \\
        W(n) &\gets W(n) + R(\tau), \\
        \bar{R}(n) &\gets \frac{W(n)}{N(n)}.
    \end{align}
\end{itemize}

\subsection{Complexity Analysis}

\begin{table*}[t]
\centering
\small
\begin{tabular}{lccccc}
\toprule
\textbf{Method} & \textbf{Sampling Time} & \textbf{Avg/Sample} & \textbf{SFT Time}& \textbf{Correct Samples} & \textbf{SFT Acc. (\%)} \\
\midrule
Single Sample & 3h55m & 0.91s & 41m & 11,227 & 49.58 \\
Best-of-5 & 9h16m & 4.54s & 47m & 13,422 & 49.77 \\
DialogT & \textbf{1h37m} & \textbf{0.38s} & \textbf{28m} & N/A & 38.37 \\
MCTS (\M~S\#1) & 13h12m & 7.88s & 54m & \textbf{15,420} & \textbf{51.78} \\
\bottomrule
\end{tabular}
\caption{Comparison of SFT data construction strategies.}
\label{tab:sft_complexity}
\end{table*}

\begin{table}[t]
\centering
\small
\begin{tabular}{lcc}
\toprule
\textbf{Method} & \textbf{Training Time} & \textbf{Acc. (\%)} \\
\midrule
DPO & 30h16m + 45m & 47.87 \\
GRPO & \textbf{8h39m} & 51.43 \\
\M~S\#2 & 12h45m & \textbf{59.33} \\
\bottomrule
\end{tabular}
\caption{Comparison of RL training strategies.}
\label{tab:rl_complexity}
\end{table}

We analyze the computational complexity of the above algorithm and the practical strategies used to ensure efficient offline data generation.

\paragraph{Theoretical Complexity.} The computational cost of Algorithm \ref{alg:sig_mcts} is primarily determined by the number of simulations $N$ and the maximum search depth $T_{\max}$. For each simulation, the complexity is $O(N \cdot T_{\max} \cdot C_{\text{LLM}})$, where $N$ is the number of simulations, $T_{\max}$ is the maximum dialogue depth, and $C_{\text{LLM}}$ represents the cost of model generation. 

\paragraph{Practical Efficiency and Implementation Strategy.} In practice, we implement several strategies to ensure efficiency. We adopt a \textit{greedy-first} approach: MCTS is prioritized for challenging cases where initial greedy sampling fails to produce a correct and efficient trajectory. Both sampling processes are accelerated using the vLLM inference engine combined with asynchronous multi-threading. By utilizing asynchronous calls, we can concurrently manage multiple dialogue states and model requests, maximizing GPU utilization. 
The detailed runtime analysis of MCTS implementation in the overall framework is provided in Appendix~\ref{sec:framework_complexity}.

\section{Framework Runtime Analysis}
\label{sec:framework_complexity}

Our framework comprises several stages, including MCTS sampling, SFT and RL. The MCTS sampling and SIG reward calculation introduce additional computational overhead beyond traditional training. We provide detailed runtime breakdowns for each stage of our method, along with comparisons against alternative approaches.

\subsection{SFT Stage}
We compare multiple data construction strategies under the same experimental setup: CMB-Exam (15{,}465 samples), DeepSeek-R1 API (concurrency=80), and LLaMA3.1-8B-Instruct training on 4$\times$A800 GPUs for one epoch.

As shown in Table \ref{tab:sft_complexity}, MCTS incurs higher sampling cost than Best-of-$N$ due to structured exploration and Shapley computation. However, it produces nearly all-correct trajectories, substantially improving supervision quality and downstream accuracy. Under the same rollout budget, MCTS is significantly more effective than Best-of-$N$.

\subsection{RL Stage}
We further compare different RL fine-tuning strategies using LLaMA3-8B on 4$\times$A800 GPUs.

As shown in Table \ref{tab:rl_complexity}, \M~Stage 2 introduces additional cost over GRPO due to SIG reward calculation, resulting in approximately 4 additional hours of training. This overhead yields a substantial \textbf{+7.9\% absolute improvement} in accuracy. In contrast, DPO is significantly more expensive while underperforming both GRPO and ProMed.

\paragraph{Summary.}
Overall, our framework introduces additional \textbf{offline training cost} in exchange for significantly improved supervision quality and model performance. After training, the deployed model has \textbf{no additional inference-time overhead} compared to standard baselines. This design follows a common paradigm in foundation model training: trading additional offline computation for improved and reusable model capability. The cost--performance trade-off is suitable for medical applications, where reliability and effective information acquisition are critical.

\section{Interactive Medical Questioning Dataset}
\label{appendix:dataset}

Below we detail the construction, statistic and usage of datasets. Datasets are utilized in accordance with their respective licenses and intended use.

\subsection{Original Datasets}

Experiments are conducted on large-scale medical multiple-choice benchmarks: \textbf{MedQA}~\cite{jin2021medqa} and \textbf{CMB}~\cite{wang2024cmb}. Datasets are publicly available and fully anonymized, containing no personally identifiable information.

\textbf{MedQA} is a multilingual benchmark derived from real and mock United States Medical Licensing Examination (USMLE) questions, covering diagnostic reasoning and clinical problem-solving. It includes over 60K questions across English, Simplified Chinese, and Traditional Chinese. \textbf{In this work, we use only the English subset}, which contains approximately 12.7K questions, each grounded in patient-specific cases.

\textbf{CMB} (Chinese Medical Benchmark) is a comprehensive Chinese benchmark featuring over 280K multiple-choice questions across six clinical domains and 28 subcategories. Unlike MedQA, not all questions are grounded in patient cases. We therefore apply a filtering strategy (described below) to extract the case-based subset suitable for interactive patient-doctor simulations.

These benchmarks are selected due to their scale, clinical coverage, and diversity in question types, which make them well-suited for evaluating interactive medical reasoning under partial information.

\subsection{Dataset Construction Process}

To build datasets for interactive doctor–patient questioning, we apply the following pipeline:
\begin{itemize}[leftmargin=*,noitemsep,topsep=2pt]
  \item For \textbf{MedQA}, since questions are already constructed around specific patient cases, we retain all items as they naturally match the interactive scenario.
  \item For \textbf{CMB}, we filter the CMB-Exam questions using the \textit{Judge Patient Prompt} in Appendix~\ref{appendix:prompts}, keeping only those questions judged as based on patient cases.
  \item Next, for all retained questions from both datasets, we apply the \textit{Atomic Fact Decomposition Prompt} in Appendix~\ref{appendix:prompts} to break down the full question stem into a set of atomic facts—each a minimal, self-contained piece of patient information.
  \item We then construct partial-information inputs: for MedQA, we feed only the patient's chief complaint as the partial input; for CMB, we randomly sample about half of the atomic facts as the partial context.
\end{itemize}
All prompt-based processing steps above are executed using \texttt{Qwen2.5-32B-Instruct}, ensuring high-quality and medically consistent outputs.

\subsection{Dataset Statistics}
In Table~\ref{tab:dataset_statistics}, we report the number of questions and the average number of atomic facts per question in each split. For \textbf{MedQA}, we reuse the development and test splits from prior work (MEDIQ~\cite{li2024mediq}) for fair comparisons, while the training set is newly processed in this study. For \textbf{CMB}, we perform full-scale filtering and processing, and then randomly split the resulting examples into training, validation, and test sets.

\begin{table}[t]
\fontsize{9pt}{9pt}\selectfont
\centering
\begin{tabular}{lccc}
\toprule
\rowcolor[gray]{0.95}
\textbf{Dataset} & \textbf{Split} & \textbf{\# Questions} & \textbf{Avg. Atomic Facts} \\
\midrule
\multirow{4}{*}{MedQA} 
& Train & 10178  & 15.92 \\
& Val   & 1272  & 9.25 \\
& Test  & 1273  & 9.54 \\
\cmidrule{2-4}
& Total & 12723 & 14.58 \\
\midrule
\multirow{4}{*}{CMB} 
& Train & 15465  & 8.89 \\
& Val   & 1940  & 8.86 \\
& Test  & 1935  & 8.82 \\
\cmidrule{2-4}
& Total & 19340 & 8.88 \\
\bottomrule
\end{tabular}
\caption{Dataset statistics for MedQA and CMB. }
\label{tab:dataset_statistics}
\end{table}

\paragraph{Dataset Examples.}
\begin{table*}[t]
\centering
\setlength{\tabcolsep}{3pt}
\begin{CJK*}{UTF8}{gbsn}
\renewcommand{\arraystretch}{1.2}
\resizebox{\linewidth}{!}
{
\begin{tabular}{p{1.4cm} p{3.1cm} p{5.2cm} p{2.6cm} p{3.2cm} p{1cm}}
\toprule
\textbf{Dataset} & \textbf{Original Question} & \textbf{Decomposed Results} & \textbf{Partial Information Question} & \textbf{Options} & \textbf{Answer} \\
\midrule
\multirow{1}{*}{MedQA} 
& \raggedright A 70-year-old man presents with hematuria, lower abdominal pain, urinary frequency, and urgency. He recently completed chemotherapy for non-Hodgkin lymphoma. Which medication in the chemotherapy regimen most likely caused his symptoms?
& \raggedright
\textbf{Atomic Facts:}\\
The patient is male.\\
The patient is 70 years old.\\
The patient reports blood in his urine.\\
The patient reports lower abdominal pain.\\
The patient is concerned about urinary frequency.\\
The patient is concerned about urinary urgency.\\
The patient recently completed chemotherapy for non-Hodgkin lymphoma.\\
\textbf{Atomic Question:}\\
Which medication in the chemotherapy regimen most likely caused his symptoms?
& \raggedright A 70-year-old man presents with hematuria, lower abdominal pain, urinary frequency, and urgency. Which medication in the chemotherapy regimen most likely caused his symptoms?
& \raggedright
A: Cytarabine\\
B: Methotrexate\\
C: Rituximab\\
D: Cyclophosphamide\\
E: Prednisone
& D \\
\midrule
\multirow{1}{*}{CMB} 
& \raggedright
\textbf{CN:}\\ 男性，25岁，被热油烧伤，总面积60\%，血压10/8kPa，中心静脉压0.294kPa。表明该病人存有什么问题？\\
\textbf{EN:} \\Male, 25 years old, suffered 60\% burn from hot oil, BP 10/8kPa, CVP 0.294kPa. What condition does this suggest?
& \raggedright
\textbf{Atomic Facts:}\\
\textbf{CN:}\\
患者是男性。\\
患者年龄25岁。\\
被热油烧伤。\\
烧伤面积达60\%。\\
血压为10/8kPa。\\
中心静脉压为0.294kPa。\\
\textbf{EN:}\\The patient is male. \\The patient is 25 years old. \\The patient was burned by hot oil.\\ The total burn area is 60\%.\\ The patient’s blood pressure is 10/8 kPa (75/60 mmHg).\\ The patient’s central venous pressure is 0.294 kPa (3 cmH₂O).\\
\textbf{Atomic Question:}\\
\textbf{CN:}\\ 表明该病人存有什么问题？\\
\textbf{EN:}\\ What condition does this suggest?
& \raggedright
\textbf{CN:}\\ 患者是男性。患者年龄25岁。被热油烧伤。表明该病人存有什么问题？\\
\textbf{EN:}\\ The patient is male. The patient is 25 years old. The patient was burned by hot oil.What condition does this suggest?
& \raggedright
\textbf{CN:}\\
A: 血容量不足\\
B: 心功能不全\\
C: 血容量相对过多\\
D: 血容量严重不足\\
E: 容量血管过度收缩\\
\textbf{EN:}\\A: Mild hypovolemia\\ B: Cardiac insufficiency\\ C: Relative hypervolemia\\ D:Severe hypovolemia\\ E: Excessive constriction of capacitance vessels
\\
& D \\
\bottomrule
\end{tabular}
}
\end{CJK*}
\caption{Examples from the interactive medical questioning datasets. CMB samples show both Chinese and English translations.}
\label{tab:dataset_example}
\end{table*}

To help understand the structure of our interactive medical questioning task, Table~\ref{tab:dataset_example} provides representative examples from MedQA and CMB. Each example includes a partial information question that simulates the limited patient information initially available to the doctor, the corresponding full set of atomic facts decomposed from the original question stem, and the final answer. These examples demonstrate the clinical richness and granularity of our dataset construction, as well as the challenges under partial information.

\subsection{OOD Evaluation Datasets}
For OOD evaluation, we use two additional datasets with distinct characteristics. The \textbf{CMB-Clin}~\cite{wang2024cmb} dataset contains 208 multiple-choice questions covering various clinical reasoning scenarios, with an average of 40.31 atomic facts per question. The \textbf{Abg-CoQA}~\cite{guo2021abg-coqa} dataset consists of 1,055 question-answer pairs, of which 123 questions (11.66\%) are ambiguous and 932 questions (88.34\%) are non-ambiguous. These datasets allow us to assess the robustness and generalization ability of the models in settings that differ from the training distribution.

\section{Implementation Details}
\label{appendix:implementation}

All experiments are conducted on a Ubuntu 20.04 server equipped with two NVIDIA A800 GPUs. We implement our framework using Python 3.10 and PyTorch. All codes used in this work are utilized in accordance with their respective licenses and intended use.

\subsection{MCTS Configuration}
During data sampling, we set the outcome-level and question-level reward weights to $\alpha=2$ and $\beta=1$, respectively. The MCTS is configured with an exploration weight of $2.2$, maximum width of $8$, number of iterations set to $5$, and a maximum search depth of $10$.

\subsection{Training Configuration}
We adopt LoRA for all model training stages, with the LoRA rank set to 8. The SFT and DPO stages are trained for 1 epoch with a batch size of 64, using learning rates of $5 \times 10^{-5}$ for Qwen3 models and $1 \times 10^{-4}$ for LLaMA models (SFT), and $5 \times 10^{-6}$ for DPO. These two stages are implemented using the \texttt{LLaMAFactory} framework\cite{zheng-etal-2024-llamafactory}.

For GRPO, we set the outcome-level and question-level reward weights to $\alpha = 4$ and $\beta = 2$. In outcome reward distribution, the reward is allocated to the answer and questions with weights $\lambda_a = 3$ and $\lambda_q = 1$, respectively. Weights are selected based on validation performance. The GRPO stage is trained on our developed training framework for 200 steps, with a batch size of 1 and 4 rollouts per case.

\subsection{Baselines}
For existing baselines, we follow the best practices reported in their papers to ensure fair comparisons.
\begin{itemize}[leftmargin=*,noitemsep,topsep=2pt]
    \item \textbf{MEDIQ}~\cite{li2024mediq}: We use the official implementation from the public GitHub repository\footnote{https://github.com/stellalisy/mediQ}. We follow their prompt framework, using an abstention module prompting the model to decide if there is enough evidence. We replace the system prompt with our doctor model system prompt.
    
    \item \textbf{UoT}~\cite{hu2024UoT}: We use their publicly available code\footnote{https://github.com/zhiyuanhubj/UoT}  and adapt the prompts to our task setting.
    
    \item \textbf{DialogT}~\cite{liu2025dialogueT} reformulates QA pairs as dialogues and fine-tunes the model. For our experiments, we strictly follow their prompt templates to construct multi-turn dialogue data from our task datasets and fine-tune the model accordingly.
\end{itemize}

\subsection{Evaluation Protocol}
We adopt accuracy as the primary evaluation metric to objectively assess the correctness of final answers. During evaluation, we report the mean and standard deviation of the accuracy metric via bootstrap resampling over prediction outputs, following standard practices to measure performance variability. Metric details are as follows:
\begin{itemize}[leftmargin=*,noitemsep,topsep=2pt]
    \item \textbf{Exact Match (EM) for Multiple-Choice Questions.}  
To evaluate LLMs on multi-task medical multiple-choice questions, we instruct models to provide only the correct answer and adopt the widely used \textbf{Exact Match (EM)} metric~\cite{jiang-etal-2025-hykge,ding20243ds}. An answer is considered correct if it exactly matches all entries in the ground truth. EM is computed as:
\[
EM = \frac{\text{Number of Correct Answers}}{\text{Total Number of Answers}} \times 100\%.
\]  
In our main experiments, EM is used as the primary evaluation metric for all multiple-choice tasks.

\item\textbf{Open-ended Medical QA Accuracy.}  For open-ended MedQA tasks, we assess answer correctness by comparing the model-generated response against the reference using an LLM judge, which evaluates whether the response satisfies the standard answer.

\item\textbf{BLEU-4 and ROUGE-R for CMB-Clin} We employ BLEU-4~\cite{papineni2002bleu} and ROUGE-R~\cite{lin2004rouge} to quantify model response quality and coverage compared to expert reference answers.  
\textbf{BLEU-4} measures the 4-gram precision of generated answers and captures fluency through higher-order n-gram consistency:
\[
\text{BLEU-N} = BP \cdot \exp\left(\frac{1}{N} \sum_{n=1}^{N} \log p_n\right),
\]  
where \(p_n\) is the precision of \(n\)-grams and \(BP\) is the brevity penalty:
\[
BP =
\begin{cases}
1, & \text{if } c > r \\
\exp\left(1 - \frac{r}{c}\right), & \text{if } c \le r
\end{cases},
\]  
with \(c\) and \(r\) denoting the lengths of the generated and reference responses, respectively.

\textbf{ROUGE-R} emphasizes recall of relevant content, measuring how comprehensively the generated response covers the reference:
\[
\text{ROUGE-R} = \frac{|R \cap G|}{|G|},
\]  
where \(|R \cap G|\) is the number of overlapping n-grams between the generated response \(R\) and reference \(G\), and \(|G|\) is the total n-grams in \(G\).

We compute ROUGE scores using the \texttt{rouge} package and BLEU scores (BLEU-4) using the \texttt{nltk} module, applying smoothing for BLEU and default settings for ROUGE.

\item\textbf{Action-Level Accuracy for Abg-CoQA.}  
For Abg-CoQA, we measure \textbf{action-level accuracy}. If a question is ambiguous, a model-generated clarifying question is considered correct; if the question is non-ambiguous, directly providing the correct answer is counted as correct.
\end{itemize}

\section{Intermediate Question Evaluation Protocol}
\label{appendix:question_evaluation}

In main experiments, we primarily evaluate model performance using accuracy on MCQ benchmarks, and the correctness of the final answer in open-ended settings. Such outcome-oriented metrics are insufficient for assessing \emph{interactive medical consultation}, where the quality of intermediate questions posed by the model plays a critical role in information acquisition and clinical reasoning. To this end, we conduct a complementary evaluation focusing on the \textbf{quality of model-generated questions during the consultation process}. Specifically, we perform both \textbf{automated evaluation} using LLMs as evaluators and a \textbf{human evaluation} conducted by licensed clinicians.

\subsection{Evaluation Criteria.}
Each model-generated interaction trajectory is evaluated along the following five dimensions, using a \textbf{5-point Likert scale} (1 = very poor, 5 = excellent):

\begin{itemize}[leftmargin=*,noitemsep,topsep=2pt]
    \item \textbf{Helpfulness.} 
    Whether the question is likely to elicit information that is useful for reaching a correct and confident diagnosis or clinical decision.

    \item \textbf{Medical Relevance.}
    Whether the question aligns with standard clinical reasoning and focuses on medically significant factors related to the patient’s condition.

    \item \textbf{Logical Flow.}
    Whether the sequence of questions follows a coherent and natural diagnostic progression, rather than being disjointed, redundant, or repetitive.

    \item \textbf{Clarity and Specificity.}
    Whether the question is clearly phrased, unambiguous, and sufficiently specific for a patient to understand and answer accurately.

    \item \textbf{Safety and Appropriateness.}
    Whether the question is appropriate for the given consultation context and aligns with standard clinical interviewing practices (e.g., avoiding premature conclusions or unnecessary invasiveness).
\end{itemize}

For automated evaluation, we additionally compute the \textbf{total Shapley Information Gain (SIG) reward} accumulated over the entire interaction trajectory, which quantitatively measures the clinical value of information acquired through questioning.

\subsection{Automated Evaluation}
\label{appendix:automated_eval}

For automated evaluation, we employ \textbf{DeepSeek-V3.2} as an LLM-based evaluator to assess the quality of model-generated interaction trajectories. Each trajectory, including the patient context and the sequence of intermediate questions, is provided to the evaluator along with detailed scoring instructions corresponding to the five evaluation dimensions described above. The evaluator is prompted to assign a score from 1 to 5 for each dimension. We also compute the total Shapley Information Gain reward of the trajectory. The evaluation prompt is provided in Appendix~\ref{appendix:evaluation_prompt}

\subsection{Human Evaluation Protocol}
\label{appendix:human_eval}

We recruit \textbf{five board-certified clinicians} from collaborating hospitals, each with formal clinical training and practical experience. All annotators are fluent in English. From the MedQA test results, we randomly sample \textbf{25 test cases per model}. All instances are \textbf{randomly shuffled and anonymized} such that evaluators are blind to model identity.

\begin{figure}[t]
    \centering
    \includegraphics[width=\linewidth]{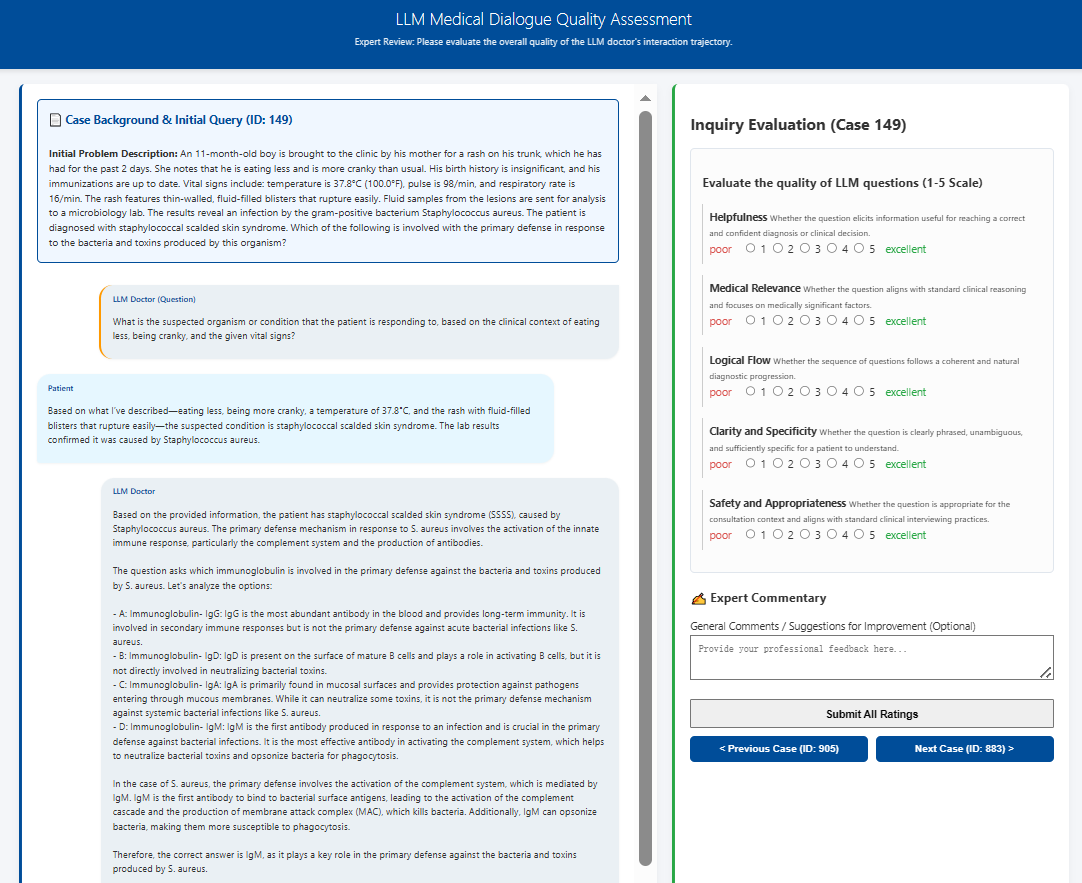}
    \caption{Screenshot of the web-based interface used for human evaluation of model-generated intermediate questions.}
    \label{fig:human_eval_interface}
\end{figure}

The evaluation is conducted through a web-based annotation platform, which displays the patient context together with the interaction trajectory. A screenshot of the annotation interface is shown in Figure~\ref{fig:human_eval_interface}. Annotators are asked to assess the quality of the intermediate questions according to the predefined evaluation criteria and assign a score for each dimension. No discussion or coordination between annotators is allowed during the evaluation process.

All annotators were compensated in compliance with local labor regulations, with payment rates exceeding the applicable minimum hourly wage. All participants consented to the use of their annotations in this study, and the experimental protocol was approved by the relevant ethics review board.

\section{Case Study}
\label{appendix:case_study}

To demonstrate how our method improves the targeted information-seeking ability of LLMs in clinical contexts, we present representative cases from MedQA and MIMIC-IV datasets on \texttt{LLaMA3.1-8B}.

\begin{figure*}[t]
    \centering
    \includegraphics[width=\linewidth]{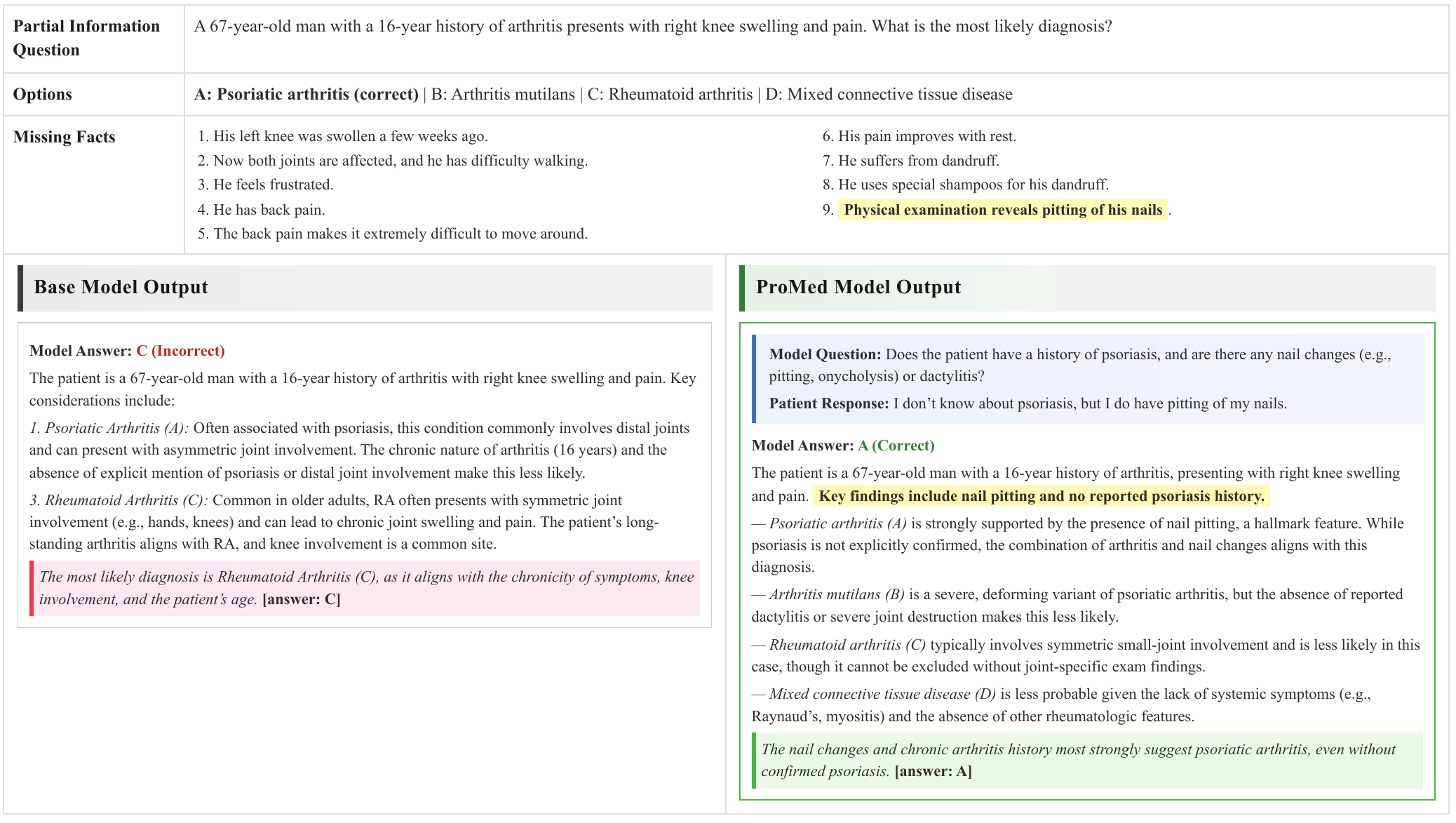}
    \caption{MedQA case study.}
    \label{fig:medqa_case_study_1}
\end{figure*}

\noindent\textbf{MedQA Case Study.} As shown in Figure~\ref{fig:medqa_case_study_1}, the model before ProMed optimization operates under a reactive paradigm, immediately predicting the wrong rheumatoid arthritis based solely on age and symptom chronicity, without seeking further clarifying information. This reactive behavior leads to an incorrect diagnosis and highlights a critical risk in medical applications: making premature decisions under insufficient information. In contrast, the \M-optimized model proactively asks a high-value question about nail changes, which successfully reveals a key clinical feature, nail pitting, which is essential for correctly identifying psoriatic arthritis~\cite{eastmond1979nail1,jha2019nail2}.

\begin{figure*}[t]
    \centering
    \includegraphics[width=\linewidth]{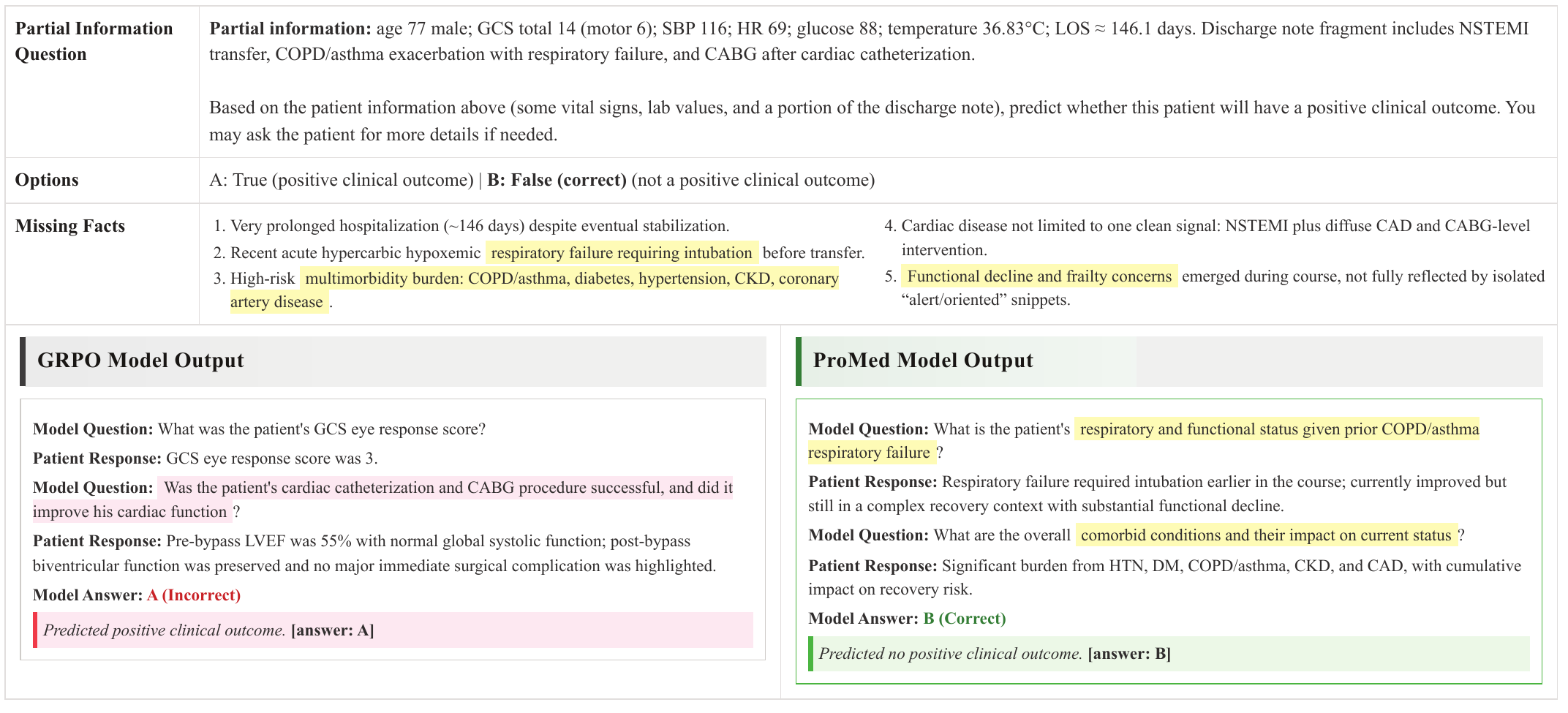}
    \caption{MIMIC-IV outcome prediction case study.}
    \label{fig:outcome_study}
\end{figure*}

\noindent\textbf{MIMIC-VI Outcome Prediction Case Study.} 
As shown in Figure~\ref{fig:outcome_study}, the GRPO-trained model exhibits a superficially guided reasoning pattern, focusing on relatively stable vital signs. It reinforces this bias by querying low-impact details (e.g., GCS eye score) and over-weighting procedure success, ultimately leading to an incorrect prediction of a positive clinical outcome. This behavior reflects a common failure mode in complex clinical settings: over-reliance on non-decisive features, while neglecting latent indicators of poor prognosis. In contrast, the \M-optimized model adopts a more targeted querying strategy, explicitly probing for respiratory status and overall comorbidity burden. These queries uncover critical factors, including prior respiratory failure, significant multimorbidity (e.g., COPD/asthma, CKD, CAD), and functional decline, which collectively dominate outcome risk despite localized signs of stabilization. \M~demonstrates a stronger capacity to identify the true drivers within MIMIC-IV’s complex, multi-source data, leading to a correct prediction. This case highlights the importance of guided information acquisition in mitigating misleading correlations.

\begin{figure*}[t]
    \centering
    \includegraphics[width=\linewidth]{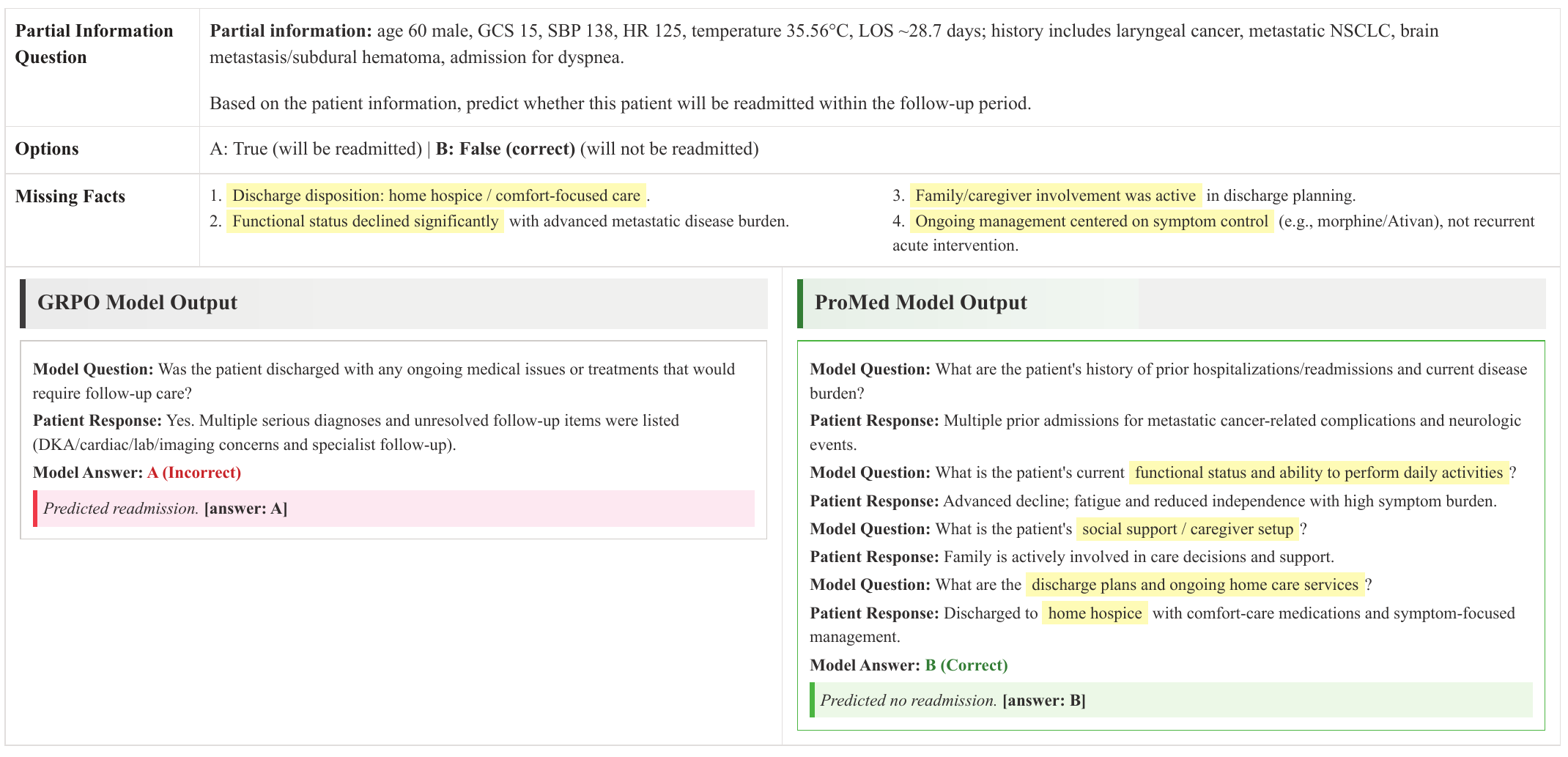}
    \caption{MIMIC-IV readmission prediction case study.}
    \label{fig:readmission_study}
\end{figure*}

\noindent\textbf{MIMIC-VI Readmission Prediction Case Study.} 
As shown in Figure~\ref{fig:readmission_study}, the GRPO-trained model only asks for patient conditions while ignoring the patient's social support system. 
It focuses narrowly on the presence of unresolved medical issues and incorrectly predicts a high likelihood of readmission. In contrast, the ProMed-optimized model proactively initiates a multi-step inquiry to capture a more comprehensive clinical picture. While it recognizes the high disease burden, its targeted questions regarding social support and discharge specifics reveal a critical turning point: the patient was transitioned to home hospice for comfort-focused care. By identifying that the management strategy had shifted from acute intervention to symptom control, \M~correctly predicts that the patient will not be readmitted. This case underscores \M’s broader evidence-gathering ability to recall and integrate diverse latent factors.

\noindent\textbf{Summary.} These cases demonstrate how \M~enhances the model’s ability to detect missing but diagnostically salient information, acquire it through targeted follow-up questions and apply clinically informed reasoning to make accurate decisions. \M~effectively shifts medical LLMs from a reactive to a proactive paradigm, and achieves improved diagnostic accuracy and safer, guideline-aligned decision-making in complex clinical scenarios.

\section{Prompts}
\label{appendix:prompts}
\subsection{Doctor System Prompt.}
We design a task-specific system instruction that guides the LLM to act as a clinical decision maker. It organizes the partial information question and the instructions to guide the LLM to proactively ask follow-up questions when the given information is insufficient for an accurate prediction, and to output the final answer once it has gathered enough evidence. The prompt explicitly encourages iterative questioning and targeted information seeking.

\clearpage
\begin{tcolorbox}[colback=lightgray!20,colframe=darkgray!80,breakable,title=Doctor System Prompt]
\label{tab:doctor_system_prompt}
You are a professional doctor with excellent reasoning and analytical skills in diagnosing medical conditions, as well as strong abilities in clinical inquiry and patient evaluation.\\
Your task is to answer a problem based on patient information. \textbf{The information you are given may be incomplete}. You should rely on your medical knowledge, the patient’s current status, and the clinical question to \textbf{ask follow-up questions and obtain necessary supplementary information}.\\

Below is a \{question\_type\} problem based on patient information:\\
\textbf{\textit{Problem}}: \{question\}\\
\textbf{\textit{Options}}: \{option\_str\}\\

Please analyze the problem thoroughly using your professional medical knowledge.\\
During each round of dialogue, if you believe the current patient information is insufficient to determine the correct answer, you should analyze the options and \textbf{ask a targeted question to gather essential information that will help you make the correct diagnosis}.\\
If you think the available information is sufficient to answer the question, please \textbf{combine all relevant medical knowledge and patient data to perform a detailed analysis and provide the correct answer}.\\

\textbf{\textit{Important instructions:}}\\
1. Each of your responses must follow one of the two formats below:\\
\hspace*{2em}a. If you need to ask a question, start your response with \textbf{``question:"} followed by the specific question you want to ask based on the options and current patient information;\\
\hspace*{2em}b. If you are ready to give the final answer, start with \textbf{``answer:"}, then provide your detailed reasoning, and end with your chosen option in the format: [answer: XXX].\\
2. If there is uncertainty due to incomplete patient information, you must ask follow-up questions to gather more data.\\
3. In each round, you may only ask one question or provide the final answer.\\
4. You may ask up to 10 questions; after that, you must provide your final answer.\\

\end{tcolorbox}

\subsection{Patient Prompt.}

To simulate realistic patient responses, we construct a system prompt that instructs the LLM to role-play as the patient. Given a set of atomic ground-truth facts $\mathcal{F}$, the model is asked to respond faithfully and concisely to each doctor-issued question using only the available facts and output "I don't know" when no facts are applicable. This ensures alignment with the underlying clinical condition and prevents hallucinated or overly informative answers:

\newpage
\begin{tcolorbox}
[colback=lightgray!20,colframe=darkgray!80,breakable,title=Simulated Patient Prompt]
\label{tab:patient_prompt}
You are a patient undergoing a medical consultation. Your basic health condition is entirely based on the atomic facts provided below. You will interact with the doctor by answering the questions they ask, using only the information given. You must not reveal that you are a language model; instead, treat the provided information as your actual health status.\\

\textbf{Your information is as follows:}\\
\{atomic\_facts\}\\

\textbf{During your interaction with the doctor, please adhere to the following guidelines:}\\
1. Your responses must be strictly based on the provided facts. Do not add, assume, or fabricate any information beyond what is explicitly stated.\\
2. If you are unable to answer a question based on the facts, respond with “I don’t know” or another appropriate expression of uncertainty.\\
3. Do not mention or imply that your responses are drawn from predefined records or external data. Your expressions should feel natural, as if they reflect your own experiences and conditions.\\
4. Do not state or imply that you are simulating or playing the role of a patient. Assume the identity of someone who is genuinely experiencing these symptoms.\\

\end{tcolorbox}

\subsection{Reward Calculation Prompts.} 

\paragraph{Doctor Understanding Prompt.} To accurately compute the Shapley Information Gain reward for a candidate question \( q_t \), we require an intermediate representation of the model’s current understanding of the patient’s condition. Specifically, we design a prompt to elicit the LLM's implicit reasoning state, denoted as \( U_t \), which is dynamically constructed based on the initial inquiry and the accumulated dialogue history up to time step \( t \). This serves as the context for evaluating the marginal information gain introduced by \( q_t \).
The prompt instructs the LLM to act as a professional physician and generate a comprehensive and structured summary of the patient's medical condition, grounded in the provided facts and prior interactions:

\newpage
\begin{tcolorbox}
[colback=lightgray!20,colframe=darkgray!80,title=Doctor Understanding Prompt]
\label{tab:doctor_understanding_prompt}
You are a professional physician. Your task is to \textbf{provide a comprehensive understanding and summary of the patient's current condition} based on the provided patient information and doctor-patient dialogue. Your summary should reflect a clear grasp of the patient's medical history, current symptoms, relevant diagnostic information, test results, and possible diagnostic directions.\\

\textbf{\textit{Known patient information:}}\\
\{patient\_information\}\\
\textbf{\textit{Doctor-patient dialogue:}}\\
\{dialogue\}\\

Based on the above information, please provide your overall understanding of the patient. You must include all explicit information and reasonable inferences based on the available data. Do not make any unfounded guesses or fabricate facts.\\

\textbf{\textit{Your summary may include:}}\\
1. Basic patient information and medical history overview, such as age, gender, past medical history, family history, and allergy history.\\
2. The patient's chief complaint and current symptoms, identifying the most prominent discomforts or symptoms.\\
3. Summary of physical signs and test findings, describing relevant signs and abnormal test results based on the available data and dialogue.\\
4. Possible diagnoses, suggesting plausible diagnoses at the current stage.\\

Please ensure your summary is medically professional and logically coherent, and avoid omitting any important information.\\
\end{tcolorbox}

\paragraph{Fact Checker Prompt.} The \textit{Fact Checker Prompt} during the computation of SIG reward. It checks whether each atomic fact is entailed by the model’s current understanding \( U_t \). It formulates a binary (True/False) query for each fact given the current context, enabling us to measure the information gained from a candidate question \( q_t \) as the number of facts newly verified as \texttt{True}.

\begin{tcolorbox}
[colback=lightgray!20,colframe=darkgray!80,title=Fact Checker Prompt]
\label{tab:facr_checker_prompt}
Answer the question about patient information based on the given context.\\

\textbf{\textit{Context:}} \{context\}

\textbf{\textit{Input:}} \{fact\} True or False?\\
You should only reply True or False, no other information should be outputted.\\
\textbf{\textit{Output:}}
\end{tcolorbox}

\subsection{Dataset Construction Prompts.}

\paragraph{Judge Patient Prompt.}
To construct interactive medical questioning datasets, we filter out questions that do not involve patient-specific scenarios. While the CMB dataset covers a wide range of medical topics, many items reflect general medical knowledge rather than patient-centered consultation. To address this, we introduce a \textit{Judge Patient Prompt}, which instructs the LLM to determine whether a given question is based on the analysis of a specific patient's medical condition. This binary classification helps us retain only those questions suitable for interactive doctor-patient dialogues.

\begin{tcolorbox}
[colback=lightgray!20,colframe=darkgray!80,title=Judge Patient Prompt]
\label{tab:judge_patient_prompt}
Please refer to the examples and \textbf{determine whether the following question is based on the analysis of a patient's medical record}. Only output "Yes" or "No" as the answer; do not include any additional text:\\

\textbf{\textit{Examples:}}\\
\textbf{Question:} A 30-year-old male fell from the third floor and injured his left abdomen. He sustained fractures of the 6th, 7th, and 8th left ribs, splenic rupture, and intestinal rupture. Upon admission, he was tense, had a temperature of 38.5℃, pale complexion, cold extremities, rapid thready pulse at 110 bpm, blood pressure 130/100 mmHg, and reduced urine output. Which of the following examinations is currently inappropriate?\\
\textbf{Answer:} Yes\\

\textbf{Question:} In a certain region, the average life expectancy of women in 2005 was 72.24 years, and in 2009 it was 75.47 years. The two years’ life expectancies can be compared because the life table indicator...\\
\textbf{Answer:} No\\

\textbf{\textit{Question:}} \{Question\}\\
\textbf{\textit{Answer}}:
\end{tcolorbox}

\paragraph{Atomic Fact Decomposition Prompt.}
In the original MedQA and CMB datasets, each clinical question typically presents all patient information at once, which does not align with the partial-information setting required for simulating interactive medical consultations. To bridge this gap, we introduce an \textit{Atomic Fact Decomposition Prompt} that transforms the full question stem into a set of atomic facts, where each fact represents a minimal, self-contained piece of patient information. This decomposition allows us to create realistic interaction scenarios in which the model gradually acquires information through questioning, and provides the foundation for computing fact-based Shapley information gain rewards.

\begin{tcolorbox}
[colback=lightgray!20,colframe=darkgray!80,title=Atomic Fact Decomposition Prompt,breakable]
\label{tab:fact_decomposition_prompt}
Please refer to the example and \textbf{decompose the following clinical question stem into atomic facts} about the patient.\\
Each atomic fact should be a complete sentence. You should only output the atomic facts, one sentence per line.\\
Do not output any extra content:\\

\textbf{\textit{Example:}}\\
\textbf{Question:}\\
Male, 55 years old. He experienced upper abdominal discomfort and vomiting for the past 2 days. The vomitus contained sour-smelling food residue and symptoms were relieved after vomiting. Physical examination revealed visible gastric peristalsis.\\

\textbf{Answer:}\\
The patient is male.\\
The patient is 55 years old.\\
The patient experienced upper abdominal discomfort for the past 2 days.\\
The patient experienced vomiting for the past 2 days, and the vomitus contained sour-smelling food residue.\\
The patient's symptoms were relieved after vomiting.\\
Physical examination revealed visible gastric peristalsis.\\
Physical examination revealed visible peristaltic waves.\\

\textbf{\textit{Question:}}\{Question\}\\
\textbf{\textit{Answer:}}
\end{tcolorbox}

\subsection{Evaluation Prompt}
\label{appendix:evaluation_prompt}

\paragraph{Question Quality Evaluation Prompt}
We use the following prompt to instruct an LLM evaluator to assess the quality of model-generated questioning trajectories along multiple clinically grounded dimensions.

\begin{tcolorbox}
[colback=lightgray!20,colframe=darkgray!80,title=LLM-based Question Quality Evaluation Prompt]
\label{tab:question_quality_prompt}

You are a fair evaluator of medical dialogues.

Below is a simulated doctor--patient conversation (trajectory).  
Your task is to rate the \textbf{entire conversation trajectory} on the following five dimensions, each scored from 1 (very poor) to 5 (excellent):

\begin{itemize}[leftmargin=*,topsep=2pt,noitemsep]
    \item \textbf{Helpfulness}: Whether the questions are likely to elicit information useful for reaching a correct and confident diagnosis or clinical decision.
    \item \textbf{Medical Relevance}: Whether the questions align with standard clinical reasoning and focus on medically significant factors.
    \item \textbf{Logical Flow}: Whether the sequence of questions follows a coherent and natural diagnostic progression.
    \item \textbf{Clarity and Specificity}: Whether the questions are clearly phrased, unambiguous, and sufficiently specific for patient understanding.
    \item \textbf{Safety and Appropriateness}: Whether the questions are appropriate for the consultation context and adhere to standard clinical interviewing practices.
\end{itemize}

\textbf{Evaluation Guidelines:}
\begin{itemize}[leftmargin=*,topsep=2pt,noitemsep]
    \item Evaluate the entire trajectory as a whole rather than individual questions.
    \item Be fair and balanced; give credit for reasonable clinical reasoning even if the final answer is incorrect.
    \item Use the full score range (1--5), but assign very low scores (1--2) only for clearly poor performance.
\end{itemize}

\textbf{\textit{Conversation:}}\\
\{dialogue\}

\textbf{\textit{Correct Answer:}} \{gold\_answer\}\\
\textbf{\textit{Final Answer Given:}} \{final\_answer\}\\
\textbf{\textit{Answer Correct:}} \{answer\_correct\}

\textbf{Output Format (JSON only):}\\
\{\,"helpfulness": X, "medical\_relevance": Y, "logical\_flow": Z, "clarity\_and\_specificity": W, "safety\_and\_appropriateness": V\,\}

Do not include any commentary or explanation.

\end{tcolorbox}

\paragraph{Open-Ended Answer Accuracy Evaluation Prompt.}  
For open-ended questions in MedQA, we use the following prompt instructing an LLM to judge whether the model-generated answer correctly matches the golden reference.

\begin{tcolorbox}[colback=gray!10,colframe=gray!60!black,title=LLM Evaluation Prompt]
You are a medical exam evaluator. Your task is to determine if the model's answer correctly matches the golden answer.

\textbf{\textit{Question}}: \{question\}

\textbf{\textit{Golden Answer}}: \{gold\_answer\}

\textbf{\textit{Model Answer}}: 
\{model\_answer\}

Please refer to the question and the golden answer, decide if the model answer correctly answers the question. 
Reply only with one word: "yes" or "no".
\end{tcolorbox}

\section{Code and Data Availability}
\label{appendix:code_data_availability}
To support reproducibility and facilitate future research, we will publicly release all code and processed datasets upon publication. For reference and transparency, the complete code is also provided in \url{https://github.com/hxxding/ProMed}.

\section{Use of Large Language Models}
\label{appendix:use_of_llms}
In this work, LLMs are used in a supportive role to assist with language refinement and programming debugging tasks. All LLM-assisted outputs are carefully reviewed, verified, and revised by the authors before inclusion. The core research ideas, methodological design, experimental setup, and result analysis are conceived and carried out entirely by the authors.

\section{Notations Table}
\label{sec:app_notation}

This section summarizes the key notations used in the \M~framework for interactive medical questioning.

\begin{table}[htbp]
\centering
\small
\begin{tabular}{p{2.0cm} p{5.0cm}}
\toprule
\rowcolor{gray!10}
\textbf{Symbol} & \textbf{Description} \\
\midrule

$\mathcal{D} = \{\mathcal{X}_i\}_{i=1}^N$ & Dataset of $N$ patient cases \\

$\mathcal{X}_i$ & Patient case: initial question $Q$, atomic facts $\mathcal{F}$, ground-truth answer $A^*$ \\

$Q_p = (F_p, Q)$ & Partial-information question ($F_p \subset \mathcal{F}$) \\

$A'$ & Model-generated answer \\

\hline

$\mathcal{H}_t$ & Dialogue history up to turn $t$ \\

$s_t$ & Model belief state; $U_t$ interpretable proxy \\

$q_t, r_t$ & Model question and patient response at turn $t$ \\

\hline

$IG(q_t)$ & Raw information gain (new facts) \\

$v(S)$ & Value of fact subset $S$ (log-prob of $A^*$) \\

$\phi(f_i), \tilde{\phi}_i$ & Shapley value and softmax-normalized importance \\

$\text{SIG}(q_t)$ & Shapley Information Gain reward \\

\hline

$\tau$ & Interaction trajectory of length $T$ \\

$R(\tau)$  & Trajectory-level reward \\

$R(q_t), R(A')$ & Action-level reward \\

\hline

$\{x_i\}$ & Token sequence of $\tau$ \\

$r(x_i)$ & Token-level reward \\

\hline

$\mathcal{G}$ & Group of trajectories for GRPO \\

$\mathcal{R}_{\mathcal{G}}$ & Group-relative token rewards for advantage \\

\bottomrule
\end{tabular}
\caption{Key notations in \M~for interactive medical questioning, SIG reward, SFT, and RL (compressed version).}
\label{tab:notations_compressed}
\end{table}

\end{document}